\documentclass{article}
\usepackage{ijcai23}

\usepackage{times}
\usepackage{soul}
\usepackage{url}
\usepackage[hidelinks]{hyperref}
\usepackage[utf8]{inputenc}
\usepackage[small]{caption}
\usepackage{graphicx}
\usepackage{amsmath}
\usepackage{amsthm}
\usepackage{booktabs}
\usepackage{algorithm}
\usepackage{algorithmic}
\usepackage[switch]{lineno}

\usepackage{amsfonts}
\usepackage{amsthm}
\usepackage{etoolbox}
\usepackage{mathtools, cuted}
\usepackage{lipsum}
\usepackage{subfigure}
\usepackage{color}
\usepackage{soul}
\usepackage{cancel}

\definecolor{mydarkgreen}{RGB}{0, 139, 69}
\hypersetup{
 colorlinks=true,
 linkcolor=red,
 urlcolor=black,
 citecolor=mydarkgreen,
}

\usepackage{bbm}
\usepackage{comment}
\usepackage{MnSymbol}

\theoremstyle{definition}
\newtheorem{exmp}{Example}

\newtheorem{theorem}{Theorem}
\newtheorem{fact}{Fact}

\newcommand{\tianping}[1]{{\color{black} #1}}
\newcommand{\zzy}[1]{{\color{black} #1}}
\title{Unbiased Gradient Boosting Decision Tree with Unbiased Feature Importance}

\author{
Zheyu Zhang \footnotemark[1]
\and
Tianping Zhang \footnotemark[1]
\And
Jian Li
\affiliations
Institute for Interdisciplinary Information Sciences (IIIS), Tsinghua University
\emails
\{zheyu-zh20, ztp18\}@mails.tsinghua.edu.cn, lijian83@mail.tsinghua.edu.cn
}

\begin{document}

\maketitle

\footnotetext[1]{Equal Contribution.}

\begin{abstract}
Gradient Boosting Decision Tree (GBDT) has achieved remarkable success in a wide variety of applications. The split finding algorithm, which determines the tree construction process, is one of the most crucial components of GBDT. However, the split finding algorithm has long been criticized for its bias towards features with a large number of potential splits. This bias introduces severe interpretability and overfitting issues in GBDT. To this end, we provide a fine-grained analysis of bias in GBDT and demonstrate that the bias originates from 1) the systematic bias in the gain estimation of each split and 2) the bias in the split finding algorithm resulting from the use of the same data to evaluate the split improvement and determine the best split. Based on the analysis, we propose {\em unbiased gain}, a new unbiased measurement of gain importance using out-of-bag samples. Moreover, we incorporate the unbiased property into the split finding algorithm and develop UnbiasedGBM to solve the overfitting issue of GBDT. We assess the performance of UnbiasedGBM and unbiased gain in a large-scale empirical study comprising 60 datasets and show that: 1) UnbiasedGBM exhibits better performance than popular GBDT implementations such as LightGBM, XGBoost, and Catboost on average on the 60 datasets and 2) unbiased gain achieves better average performance in feature selection than popular feature importance methods. The codes are available at \url{https://github.com/ZheyuAqaZhang/UnbiasedGBM}.
\end{abstract}

\section{Introduction}
Gradient Boosting Decision Tree (GBDT) is 
one of the most widely used machine learning models and has been applied in numerous domains, including medicine, finance, climate science, and healthcare. GBDT is particularly popular in modeling tabular data~\cite{xgboost,deeplearning-not,revisiting,why-tree}. In the training process of a GBDT model, we need to construct decision trees one by one. During tree construction, we need to determine the best split and the split finding algorithm is one of the most crucial components in GBDT.
In the standard implementation \cite{friedman2001greedy,xgboost,lightgbm}, the best split is chosen based on the reduction in loss (decrease in impurity) of all candidate splits in all features. However, this split finding algorithm has long been criticized for its bias towards features that exhibit more potential splits~\cite{breiman2017classification,strobl-bias,boulesteix-bias,nicodemus2011stability}. Due to the increased flexibility afforded by a larger number of potential split points, features with higher cardinality (such as continuous features and features with a large number of categories) have a higher probability of being split than features with lower cardinality (such as binary features). This bias introduces two problems in GBDT:

\begin{itemize}
    \item Interpretability issue. The \tianping{{\em gain importance}~\cite{breiman2017classification,DBLP:books/sp/ESL}} in GBDT sums up the total reduction of loss in all splits for a given feature, and is frequently used to explain how influential a feature is on the models' predictions. However, gain importance is not reliable due to its bias towards features with high cardinality~\cite{breiman2017classification}. \tianping{As we illustrate in Example~\ref{example: X123} in Section 4, a continuous feature independent of the target may have higher gain importance than a binary feature related to the target.} 
    \item Overfitting issue. During tree construction, the split finding algorithm biases towards choosing features with high cardinality\tianping{~\cite{breiman2017classification,strobl-bias}}. Moreover, the split finding algorithm uses training set statistics to determine the best split and does not evaluate the generalization performance of each split.
\end{itemize}

Existing studies to address the bias problems mostly fall into two categories: 1) they propose a post hoc approach to calculate unbiased or debiased feature importance measurement~\cite{unbiased,debiased}, and 2) they propose new tree building algorithms by redesigning split finding algorithms~\cite{QUEST,CRUISE,GUIDE,strobl-bias}. However, these methods mostly focus on random forests, and cannot generalize to GBDT. \tianping{One of the main reasons is that, different from most random forest implementations, existing GBDT implementations employ the second-order approximation of the objective function to evaluate split-improvement~\cite{xgboost,lightgbm}} (see more detailed discussions in the related work section). Since popular GBDT implementations, such as XGBoost~\cite{xgboost}
and CatBoost~\cite{catboost}, have been dominating tabular data modeling~\cite{revisiting,deeplearning-not}, there is an urgent need to address the interpretability and overfitting issues caused by the bias in GBDT.

To study the causes of the bias in GBDT, we conduct a fine-grained analysis, which reveals that the bias originates from: 1) the systematic bias in each split's gain estimation. We discover that the calculation of gain is a biased estimation of the split improvement, and is almost always positive. 2) The bias in the split finding algorithm due to the fact that it evaluates the split improvement and determines the best split using the same set of data. According to the analysis, first, we construct an unbiased measurement of feature importance for GBDT by using out-of-bag samples. This new measurement is unbiased in the sense that features with no predictive power for the target variable has an importance score of zero in expectation. Next, we incorporate the unbiased property into the split finding algorithm during tree construction and propose UnbiasedGBM. \tianping{Compared with existing GBDT implementations (such as LightGBM~\cite{lightgbm}, XGBoost~\cite{xgboost}, and CatBoost~\cite{catboost}), UnbiasedGBM has two advantages}:
\begin{enumerate}
    \item The split finding algorithm unbiasedly chooses among features with different cardinality to mitigate overfitting.
    \item UnbiasedGBM evaluates the generalization performance of each split and performs leaf-wise early-stopping to avoid overfitting splits.
\end{enumerate}
The contributions of this paper are summarized as follows:

\tianping{
\begin{enumerate}
    \item We propose unbiased gain, an unbiased measurement of feature importance in GBDT to address the interpretability issue due to the bias in the split finding algorithm.
    \item We propose UnbiasedGBM by integrating the unbiased property into the split finding algorithm to mitigate overfitting. 
    \item We provide a large-scale empirical study comprising 60 datasets to show that: 1) UnbiasedGBM exhibits better performance on average than LightGBM, XGBoost, and Catboost, and 2) unbiased gain achieves better average performance in feature selection than gain importance, permutation feature importance, and SHAP importance.
\end{enumerate}
}

\section{Related Work}
Existing methods to correct the bias in the split finding algorithm fall primarily into two categories: 1) they propose a new method to compute debiased or unbiased feature importance measurement. 2) They propose new tree construction algorithms by redesigning the split finding algorithm.

There has been a line of work to develop new methods for computing debiased or unbiased feature importance. Quinlan~\cite{DBLP:journals/ml/Quinlan86} proposed information gain ratio to overcome the bias in classification trees. Sandri and Zuccolotto~\cite{sandri2008bias} decomposed split-improvement into the reduction in loss and a positive bias. They used a pseudo dataset to estimate and subtract the bias. Nembrini et al.~\cite{nembrini2018revival} then improved the computing efficiency of this approach. Li et al.~\cite{debiased} proposed a debiased feature importance measure. However, their method still yields biased results. Zhou and Hooker~\cite{unbiased} proposed an unbiased measurement of feature importance in random forests. Nonetheless, the theoretical analysis relies on using mean squared error to justify the unbiased property of their method and cannot be generalized to GBDT, which often employs different loss functions for tree construction. In this paper, we propose unbiased gain, an unbiased measurement of feature importance in GBDT. Our method enjoys several advantages compared with previous methods: 1) Our method does not generate pseudo data that incurs additional cost as in Sandri and Zuccolotto~\cite{sandri2008bias} and Nembrini et al.~\cite{nembrini2018revival}. 2) Our method can be easily used in GBDT implementations and has the theoretical guarantee of being unbiased, whereas Zhou and Hooker~\cite{unbiased} cannot generalize to GBDT.

There has been another line of works that develop new tree building algorithms to remove the bias, such as QUEST~\cite{QUEST}, CRUISE~\cite{CRUISE}, GUIDE~\cite{GUIDE}, and cforest~\cite{strobl-bias}. However, these methods cannot generalize to GBDT for a variety of reasons. For example, QUEST, CRUISE, and GUIDE use classification trees, whereas GBDT uses regression trees for both classification and regression tasks and supports various loss functions. cforest~\cite{strobl-bias} separates the variable selection and the splitting procedure to remove the bias. However, this method incurs an excessive amount of computational overhead, as variable selection is typically costly. We are the first to integrate the unbiased property into GBDT and develop UnbiasedGBM to address the overfitting problem caused by the bias in GBDT.

\section{Background}
We briefly introduce the GBDT model
in which the second order approximation is used in the training 
(e.g., XGBoost \cite{xgboost}, LightGBM \cite{lightgbm}).
Note that we formulate the GBDT objective under the population distribution as opposed to the traditional formulation of GBDT utilizing the empirical distribution~\cite{xgboost,li2012robust}. This formulation is essential, and it allows us to examine and comprehend the bias in GBDT.
\subsection{Gradient Boosting Decision Trees}

Consider the dataset $D=\{(\mathbf{x},y)\}$, where $(\mathbf{x},y)$ are independent and identically distributed from an unknown distribution $\mathcal{T}$. A tree ensemble model uses $K$ additive functions to model the distribution $\mathcal{T}$ and predict the output:
\begin{linenomath}
\[
\hat y = \phi(\mathbf x) = \sum_{k=1}^{K} f_k(\mathbf{x}), f_k \in \mathcal F,
\]
\end{linenomath}
where $\mathcal F = \{f(\mathbf x) = w_{q(\mathbf x)}\}$ is the space of regression trees. Here $q$ represents the tree structure and $q(\mathbf{x})$ maps an example $\mathbf{x}$ to the leaf index. We construct the tree ensemble in an additive manner to minimize the objective function $\mathcal L(\phi) = \mathbb E_{\mathbf x, y} \left[ l(\phi(\mathbf{x}), y) \right].$ Let $\phi_{t}$ be the model at the $t$-th iteration and $\hat{y}_t$ be the corresponding prediction. We greedily add a new regression tree $f_t$ that most improves the objective function $\mathcal{L}(\phi_{t-1}+f_t)$. This is achieved by using the second-order approximation:
\begin{linenomath}
\begin{align*}
    \mathcal{L}(\phi_{t-1}+f_t) \approx \mathbb E_{\mathbf x, y}&[l(\hat{y}_{t-1},y)+g(\mathbf{x},y)f_t(\mathbf{x})\\
    &+\frac{1}{2}h(\mathbf{x},y)f_t(\mathbf{x})^2] ,
\end{align*}
\end{linenomath}
where
\begin{linenomath}
$$
g(\mathbf{x},y) = \frac{\partial l(\phi_{t-1}(\mathbf x),y)}{\partial \phi_{t-1}(\mathbf x)},\ 
h(\mathbf{x},y) = \frac{\partial^2 l(\phi_{t-1}(\mathbf x),y)}{\left(\partial \phi_{t-1}(\mathbf x) \right)^2}.
$$
\end{linenomath}
We can simplify the objective function by removing the constant terms:
\begin{linenomath}
$$
\Tilde{\mathcal{L}}(\phi_{t-1}+f_t) = \mathbb E_{\mathbf x, y}\left[g(\mathbf{x},y)f_t(\mathbf{x})+\frac{1}{2}h(\mathbf{x},y)f_t(\mathbf{x})^2\right].
$$
\end{linenomath}
For a leaf node $I$ in the tree structure, the loss $\mathcal{L}(I)$ contributed by the leaf is
\begin{linenomath}
\begin{align*}
    \mathcal L(I) &= \mathbb E_{\mathbf x, y} \left[ \mathfrak{1}_{\{q(\mathbf{x})=I\}}\left(g(\mathbf x,y) f(x) + \frac{1}{2} h(\mathbf x,y) f(x)^2 \right) \right] \\
    &= \mathbb E_{\mathbf x, y} \left[ \mathfrak{1}_{\{q(\mathbf{x})=I\}}\left(g(\mathbf x,y) w_I + \frac{1}{2} h(\mathbf x,y) w_I^2 \right) \right] \\
    &= P(\mathbf{x} \in I) \left(\mu_g(I)w_I + \frac{1}{2} \mu_h(I) w_I^2\right),
\end{align*}
\end{linenomath}
where $\mu_g(I)=\mathbb E_{\mathbf x, y}\left[g(\mathbf{x},y) \right]$ and $\mu_h(I)=\mathbb E_{\mathbf x, y}\left[h(\mathbf{x},y) \right]$. We can calculate the optimal weight $w_I$ of leaf $I$ by
\begin{linenomath}
$$
w_I=-\frac{\mu_g(I)}{\mu_h(I)}
$$
\end{linenomath}
and compute the corresponding optimal loss by
\begin{linenomath}
\begin{equation}
\label{equ: node loss}
\mathcal L(I) = - \frac{1}{2} \frac{\mu_g(I)^2}{\mu_h(I)} P(\mathbf{x} \in I).
\end{equation}
\end{linenomath}

Consider a split on feature $X_j$ at a splitting point $s$, which results in two child nodes $I_L = \left\{(\mathbf x,y)|x_j \le s\right\}\ \text{and}\ I_R = \left\{(\mathbf x,y)|x_j > s\right\}$. The gain of the split $\theta=(j,s)$ is defined as the reduction in loss:
\begin{linenomath}
\begin{equation}
\label{equ: gain}
\mathrm{Gain}(I,\theta) = \mathcal L(I) - \mathcal L(I_L) - \mathcal L(I_R).
\end{equation}
\end{linenomath}

In practice, the distribution $\mathcal{T}$ is usually unknown, therefore we cannot directly calculate $\mu_g(I)$ and $\mu_h(I)$. Instead, we use the training dataset to estimate $\mu_g(I)$ and $\mu_h(I)$. Given a training dataset with $n$ examples and $m$ features $\mathcal D_{tr}=(X,Y)=\{(\mathbf{x}_i,y_i)\}$, where $|\mathcal D_{tr}|=n$ and $(\mathbf{x}_i, y_i)\stackrel{\mathrm{iid}}{\sim}\mathcal{T}$, we estimate the loss on leaf $I$ and the gain of a split by
\begin{linenomath}
\begin{align}
    \widetilde {\mathcal L}(I) &= -\frac{1}{2} \frac{\left( \frac{1}{n_I} \sum_{i\in I} g_i\right)^2}{\frac{1}{n_I} \sum_{i\in I} h_i} \frac{n_I}{n} = -\frac{1}{2n} \frac{G_I^2}{H_I}, \label{eq: widetilde L}\\
    \widetilde {\mathrm{Gain}}(I,\theta) &= \frac{1}{2n} \left(\frac{G_L^2}{H_L} +\frac{G_R^2}{H_R} - \frac{G_I^2}{H_I} \right), \label{eq: split_gain}
\end{align}
\end{linenomath}
where $G_I=\sum_{i\in I}g_i$, $H_I=\sum_{i\in I}h_i$, and $n_I$ is the number of samples on node $I$.

\subsection{Gain Importance}
Gain importance~\cite{breiman2017classification,DBLP:books/sp/ESL}, also known as mean decrease in impurity, is a kind of feature importance in tree-based methods. It is frequently used to explain how influential a feature is on the model's predictions. Gain importance is calculated by summing up the split gain in Eq \ref{eq: split_gain} of all the splits for each feature respectively.

\section{Analysis of Bias in GBDT}
We analyze the bias in GBDT and demonstrate that it stems from the systematic bias in the gain estimation and the bias in the split finding algorithm.
We show how this bias might lead to serious interpretability and overfitting problems in GBDT.

\subsection{Bias in The Gain Estimation}
$\widetilde {\mathrm{Gain}}$ estimates the reduction in loss of a given split on a feature, which is used in both tree construction and the interpretation of a feature's importance in the tree ensemble model. Intuitively, we would like the $\widetilde {\mathrm{Gain}}$ to be unbiased, i.e., it should be zero in expectation when randomly splitting on a feature that is independent of the target. However, $\widetilde {\mathrm{Gain}}$ is always non-negative for any split on any feature.

\begin{theorem} \label{thm: gain ge zero}
For a dataset $(X, Y)$ sampled from a distribution $\mathcal{T}$, for any split $\theta$ of node $I$ on a given feature $X_j$, we always have
\begin{linenomath}
$$ \widetilde {\mathrm{Gain}}(I,\theta)\geq 0.$$
\end{linenomath}
\end{theorem}
According to the theorem, the split gain for a random split on a feature independent of the target is almost always positive (the split gain is zero in very rare cases, see the proof and more discussions in Appendix \ref{app: sec: proof 1}). This implies that 1) we may split on an uninformative feature, and 2) a positive split gain does not necessarily indicate that the feature contributes to the model. One of the reasons causing this bias is that, $(\frac{1}{n}\sum_{i=1}^n g_i)^2$ is not an unbiased estimation of $\mu_g^2$:
\begin{linenomath}
\begin{align*}
 &\ \mathbb E_{\mathcal D}\left[ \left(\frac{1}{n_I} \sum_{i\in I} g_i\right)^2 \right] \\
    = &\ \mathbb E_{\mathcal D}\left[\frac{1}{n_I^2} \sum_{i,j\in I, i\ne j} 2g_i g_j \right] + \mathbb E_{\mathcal D}\left[\frac{1}{n_I^2} \sum_{i\in I} g_i^2 \right] \\
    = &\ \frac{n_I-1}{n_I} \mu_g(I)^2 + \frac{1}{n_I} \left(\mu_g(I)^2 + \sigma_g(I)^2\right) \\
    = &\ \mu_g(I)^2 + \frac{1}{n_I} \sigma_g(I)^2.
\end{align*}
\end{linenomath}
For an uninformative feature that is independent of the target, any split on the feature yields $\mu_g(I)=\mu_g(I_L)=\mu_g(I_R)$ and $\sigma_g(I)=\sigma_g(I_L)=\sigma_g(I_R)$. Consider a regression problem with the MSE loss where the hessian is always a constant, according to Eq. \ref{eq: widetilde L} we have
\begin{linenomath}
$$\mathbb E_{\mathcal D}\left[\widetilde {\mathcal L}(I)\right] = \frac{1}{2n}\left( \mathbb E_{\mathcal D}\left[n_I\right] \mu_g(I)^2 + \sigma_g(I)^2 \right),$$
\end{linenomath}
hence the split gain on the uninformative feature is
\begin{linenomath}
\begin{align*}
    & \mathbb E_{\mathcal D} \left[ \widetilde {\mathrm{Gain}}(I, \theta)\right] \\
    =\ & \mathbb E_{\mathcal D}\left[\widetilde {\mathcal L}(I_L)\right] + \mathbb E_{\mathcal D}\left[\widetilde {\mathcal L}(I_R)\right] - \mathbb E_{\mathcal D}\left[\widetilde {\mathcal L}(I)\right]  \\
    =\ & \frac{1}{2n} \left(\sigma_g(I)^2 +  \mathbb E_{\mathcal D}\left[{n_L+n_R-n_I}\right]\mu_g(I)^2 \right) \\
    =\ & \frac{1}{2n} \sigma_g(I)^2 \geq 0.
\end{align*}
\end{linenomath}

\subsection{Bias in The Split Finding Algorithm} \label{sec:potentialsplitsbias}

One of the main challenges in tree learning is to find the optimal split that maximize the reduction of the loss, as shown in Eq. \ref{equ: gain}. To do this, a split finding algorithm iterates over candidate splits on all features to identify the optimal split that minimizes loss on the training dataset. This strategy for identifying the optimal split introduces two problems in tree learning: 1) the split finding algorithm favors features with high cardinality (such as continuous features or categorical features with many categories). Higher cardinality features have a greater number of candidate splits, and thus a greater likelihood of being split. 2) The split finding algorithm always selects the best split on the training set, without evaluating the generalization performance of each split. The two problems together lead to the overfitting problem in GBDT. We use an example to illustrate how these two problems adversely affect tree learning.

\begin{figure}[t]
\centering
    \subfigure[Gain importance of GBDT]{ 
        \begin{minipage}[t]{0.48\linewidth}
        \centering
        \includegraphics[width=\textwidth]{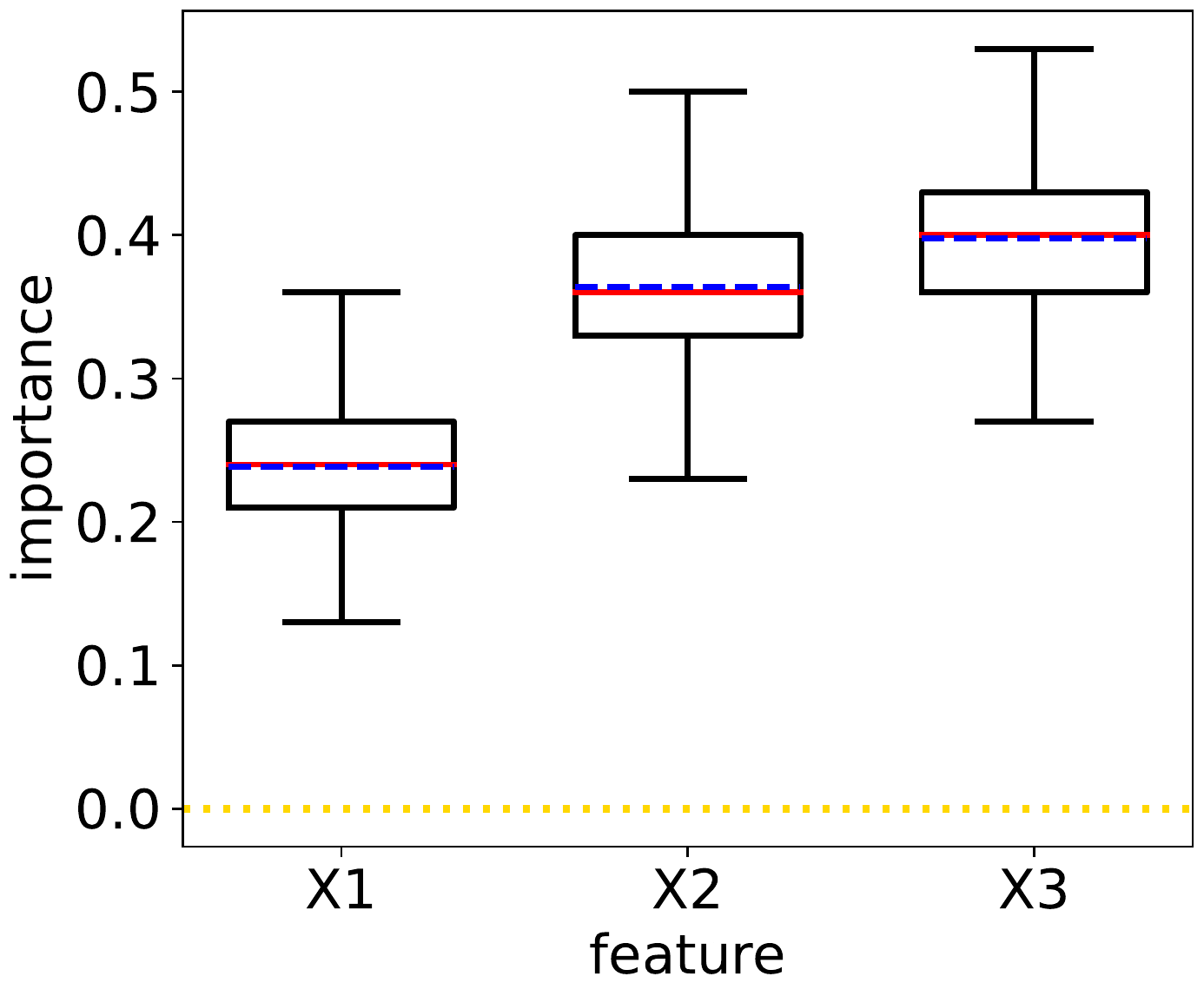}
        \label{fig: toy-1}
        \end{minipage}%
    }
    \subfigure[Unbiased gain of GBDT]{
        \begin{minipage}[t]{0.48\linewidth}
        \centering
        \includegraphics[width=\textwidth]{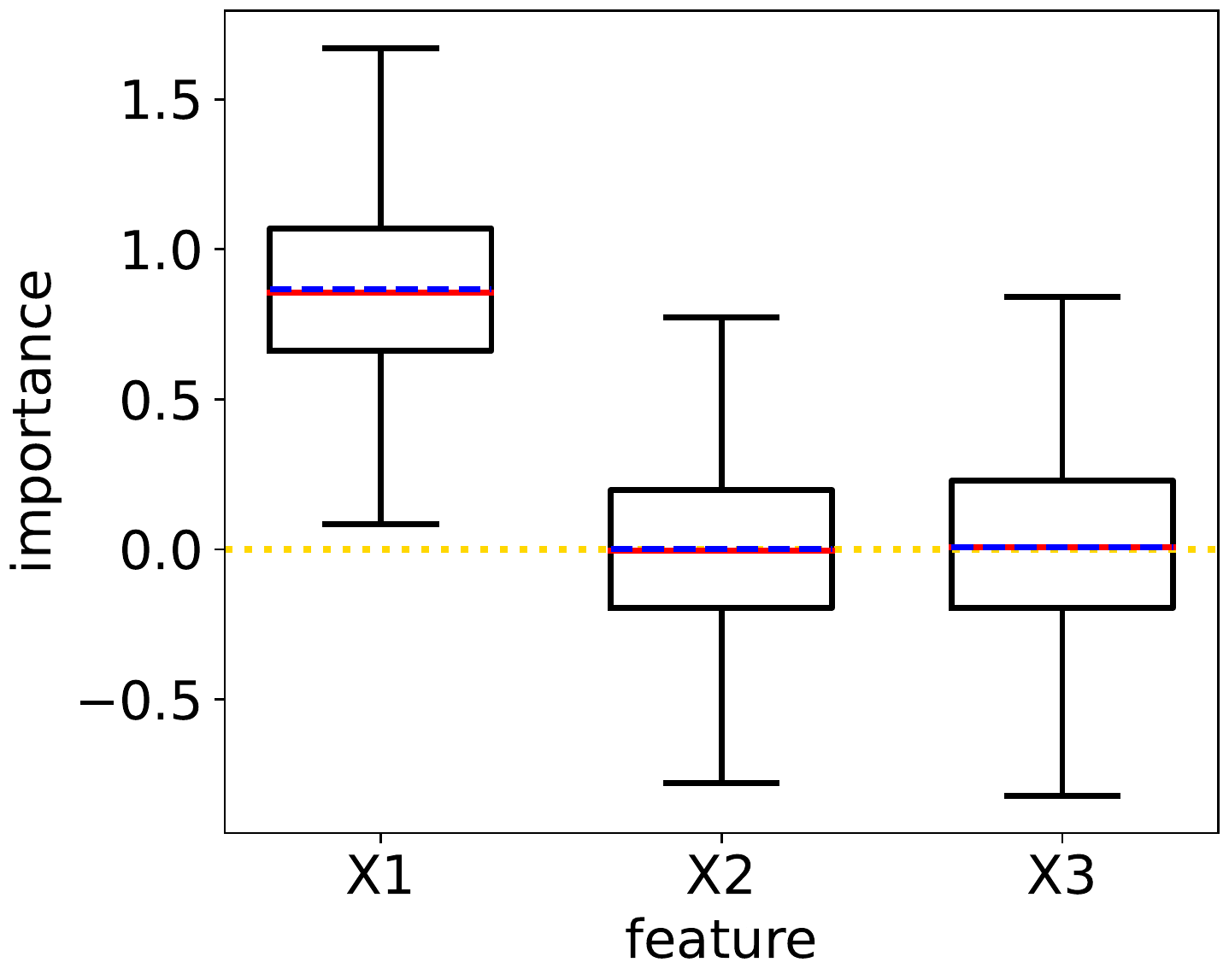}
        \label{fig: toy-2}
        \end{minipage}%
    }
\caption{The features' importance under different importance measurements in the synthetic dataset. The box plot is based on 1000 repetitions. \tianping{Unbiased gain correctly assigns the highest importance to $X_1$ and an importance of zero in expectation to $X_2$ and $X_3$.}}
\label{fig:toy}
\end{figure}

\begin{exmp} \label{example: X123}
We generate a synthetic dataset, so that $X_1$ is a binary feature, $X_2$ is a categorical feature with 6 categories (each category has equal probability), and $X_3\sim N(0,1)$ is continuous. Consider a regression problem with $y=0.1 X_1+\epsilon$ where $\epsilon\sim N(0,1)$. We train a GBDT on the synthetic dataset and plot the gain importance of each feature in Figure~\ref{fig: toy-1}. We can see that the importance of $X_2$ and $X_3$ is larger than that of $X_1$, even if $X_2$ and $X_3$ are independent with the target variable. This shows that GBDT overfits on the noise due to the bias in the split finding algorithm. In addition, this bias introduces interpretability issues, as $X_2$ and $X_3$ are more important than $X_1$ based on the gain importance.
\end{exmp}

\section{Our Method}
To solve the interpretability issue caused by the bias in GBDT, we propose ``unbiased gain", an unbiased measurement of feature importance in Section~\ref{sec: unbiased gain}. Then we incorporate the unbiased property into the split finding algorithm and propose UnbiasedGBM in Section~\ref{sec: unbiasedGBM} to address the issue of overfitting caused by the bias in GBDT.

\subsection{Unbiased Gain}
\label{sec: unbiased gain}
Our earlier analysis revealed that there are two sources of the bias in gain importance. First, gain importance biases towards features with high cardinality due to the split finding algorithm. Second, gain importance is always non-negative due to the biased estimation of Eq \ref{equ: gain}. Our goal is to propose an unbiased measurement of feature importance (unbiased gain). This new measurement is unbiased in a sense that an uninformative feature will receive an importance score of zero in expectation. 

In order to design an unbiased measurement of feature importance, we need to eliminate two sources of bias in the current gain importance measurement mentioned above. The intuitive rationale for the first source of bias is that we should not determine the best split and assess its performance using the same set of data. Therefore, a separate validation set is considered to estimate the gain importance. However, directly computing gain importance using the validation set still suffers from the second source of bias. Therefore, we construct a new form of estimation using the validation set that meets the zero-expectation criterion.

Assume we have a training dataset $\mathcal D=\{(x_i,y_i)\}$ and a validation dataset $\mathcal D'=\{(x_i',y_i')\}$. For a given leaf node $I$ and a given split $I=I_L\cup I_R$, there are $n_I$, $n_L$, $n_R$ training examples and $n_I'$, $n_L'$, $n_R'$ validation examples that fall into leaf node $I$, $I_L$, and $I_R$. First, we estimate $\mu_g(I)$ using the training examples
\begin{linenomath}
$$
\hat{\mu}_g(I)=\frac{1}{n_I}G_I=\frac{1}{n_I}\sum_{q(x_i)=I}g_i.
$$
\end{linenomath}

Then, we randomly select $k$ examples from $n_I'$ validation examples, where $k=\min(n_L', n_R')$. Next, we estimate $\mu_g(I)$ and $\mu_h(I)$ using $k$ randomly selected validation examples
\begin{linenomath}
\begin{align*}
    \hat{\mu}_g'(I)=& \frac{1}{k}G_I'=\frac{1}{k}\sum_{q(x_i')=I}g_i'\cdot \delta(I,i), \\
    \hat{\mu}_h'(I)=& \frac{1}{k}H_I'=\frac{1}{k}\sum_{q(x_i')=I}h_i'\cdot \delta(I,i),
\end{align*}
\end{linenomath}
where $\delta(I,i)$ is a binary indicator showing whether a validation sample has been selected. Finally we can calculate the loss of leaf node $I$ by

\begin{linenomath}
$$
\widetilde{\mathcal{L}}(I)=\frac{1}{2}\frac{\hat{\mu}_g(I)\cdot \hat{\mu}_g'(I)}{\hat{\mu}_h'(I)}\cdot \frac{n_I}{n}=-\frac{1}{2n}G_I\cdot \frac{G_I'}{H_I'}\ .
$$
\end{linenomath}

Here, $G_I$ is computed using the training set while $G_I'$ and $H_I'$ are computed using the validation set. We can also calculate $\widetilde{\mathcal{L}}(I_L)$ and $\widetilde{\mathcal{L}}(I_R)$ in a similar way (the number of selected validation example $k$ is the same for $I$, $I_L$, and $I_R$). Finally, the unbiased gain is calculated as
\begin{linenomath}
\begin{equation}
\label{eqn: unbiased gain}
    \widetilde{\mathrm{Gain}}_{\mathrm{ub}}(I,\theta) = 
      \widetilde{\mathcal{L}}(I) -
     \widetilde{\mathcal{L}}(I_L) - \widetilde{\mathcal{L}}(I_R).
\end{equation}
\end{linenomath}
\tianping{
\begin{theorem}
For a feature $X_j$, a leaf node $I$, and \zzy{a split $\theta$}, if $X_j$ is marginally independent of $y$ within the region defined by the leaf node $I$, then
\begin{linenomath}
$$\mathbb E_{\mathcal D'} \left[ \widetilde{\mathrm{Gain}}_{\mathrm{ub}}(I,\theta) \right] = 0.$$
\end{linenomath}
\end{theorem}}

A critical design in the unbiased gain is that, instead of estimating $\mu_g(I)$, $\mu_g(I_L)$, and $\mu_g(I_R)$ using all the validation examples on node $I$, $I_L$, and $I_R$, we randomly select $k$ examples from node $I$, $I_L$, and $I_R$ respectively for estimation. This design is critical for the unbiased property of Eq \ref{eqn: unbiased gain} (see the proof of Theorem 2 and more explanations in Appendix~\ref{app: sec: proof})

The unbiased gain we propose serves as a post hoc method to address the interpretability issue. In Figure \ref{fig: toy-2}, we plot the unbiased gain of the GBDT trained on the synthetic data. We can see that the unbiased gain correctly assigns $X_1$ with the highest importance, and the importance of $X_2$ and $X_3$ is zero in expectation.

\subsection{UnbiasedGBM} 
\label{sec: unbiasedGBM}
We propose UnbiasedGBM to address the overfitting problem introduced by the bias in GBDT: 1) The choice of each split biases towards features with high cardinality. 2) We always choose the best split on the training set, without evaluating the generalization performance of each split.

In order to eliminate these two biases, we need two validation sets. Assume we divide the training set into a sub-training set $\mathcal D$ and two validation sets $\mathcal D_1'$ and $\mathcal D_2'$. UnbiasedGBM eliminates the bias by redesigning the split finding algorithm. The design is conceptually simple but requires a good understanding of the bias in GBDT. First, we calculate the gain of each split $\widetilde{\mathrm{Gain}}_1$ in the original fashion using the sub-training set $\mathcal D$. We determine the best split of each feature using $\widetilde{\mathrm{Gain}}_1$ of each split. Next, we calculate the gain $\widetilde{\mathrm{Gain}}_2$ of each feature's best split using the validation set $\mathcal D_1'$. We determine which feature to split using $\widetilde{\mathrm{Gain}}_2$ of each feature's best split. Since we determine the best split of each feature and the feature to split using different data, we only need to consider the best split of each feature when choosing the feature to split, thus eliminating the bias towards features with high cardinality. Finally, we use the data set $\mathcal D_2'$ to calculate the unbiased gain $\widetilde{\mathrm{Gain}}_{\mathrm{ub}}$ of the best split. $\widetilde{\mathrm{Gain}}_{\mathrm{ub}}$ measures the generalization performance of the best split. We split on the leaf node if $\widetilde{\mathrm{Gain}}_{\mathrm{ub}} > 0$ and stop if $\widetilde{\mathrm{Gain}}_{\mathrm{ub}}\leq 0$.

\textbf{Remark}. We perform early-stopping on a leaf node when the best split has $\widetilde{\mathrm{Gain}}_{\mathrm{ub}}\leq 0$. However, this negative $\widetilde{\mathrm{Gain}}_{\mathrm{ub}}$ is taken into account when computing the importance of each feature in UnbiasedGBM to maintain the unbiased property.

To sum up, UnbiasedGBM enjoys two advantages over the existing GBDT: 1) UnbiasedGBM unbiasedly chooses among features with different cardinality to mitigate overfitting. 2) UnbiasedGBM measures the generalization performance of each split and performs leaf-wise early-stopping to avoid overfitting splits.

\textbf{Discussion.} Existing GBDT implementations can also perform leaf-wise early-stopping by using the minimal gain to split. However, this method and our method have two conceptual differences. First, we measure the generalization performance of each split, whereas existing methods only use statistics on the training set. Second, our ``minimal gain to split'' is zero on a theoretic basis, whereas existing methods require heuristic tuning of the minimal gain to split.

\textbf{Implementation details.} An important detail is how to divide the dataset into $\mathcal D$, $\mathcal D_1'$, and $\mathcal D_2'$. We experiment with different ratios of splitting the dataset and find out that we achieve the best performance when $|\mathcal D|=|\mathcal D_1'|=|\mathcal D_2'|$ (see more details in Appendix~\ref{app: sec: ImpDetail}).
An intuitive explanation is that different datasets are equally important in our algorithm and should have the same number of samples.

\section{Experiments}

 \begin{figure*}[ht]
\centering
    \subfigure[20 small-scale datasets with only numerical features.]{
        \begin{minipage}[t]{0.48\linewidth}
        \centering
        \includegraphics[width=\textwidth]{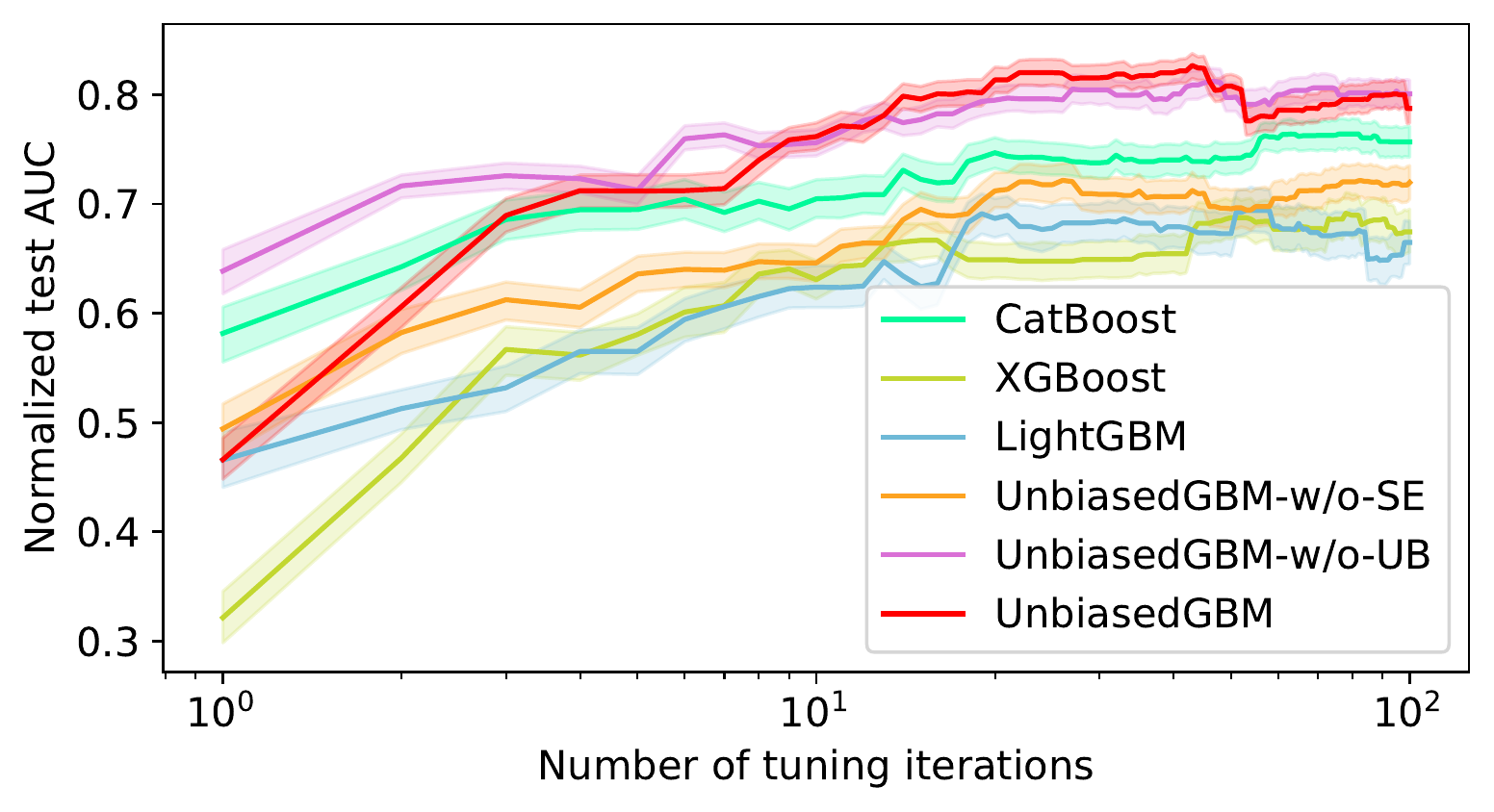}
        \end{minipage}%
    }%
    \subfigure[20 small-scale datasets with num. and cat. features.]{
        \begin{minipage}[t]{0.48\linewidth}
        \centering
        \includegraphics[width=\textwidth]{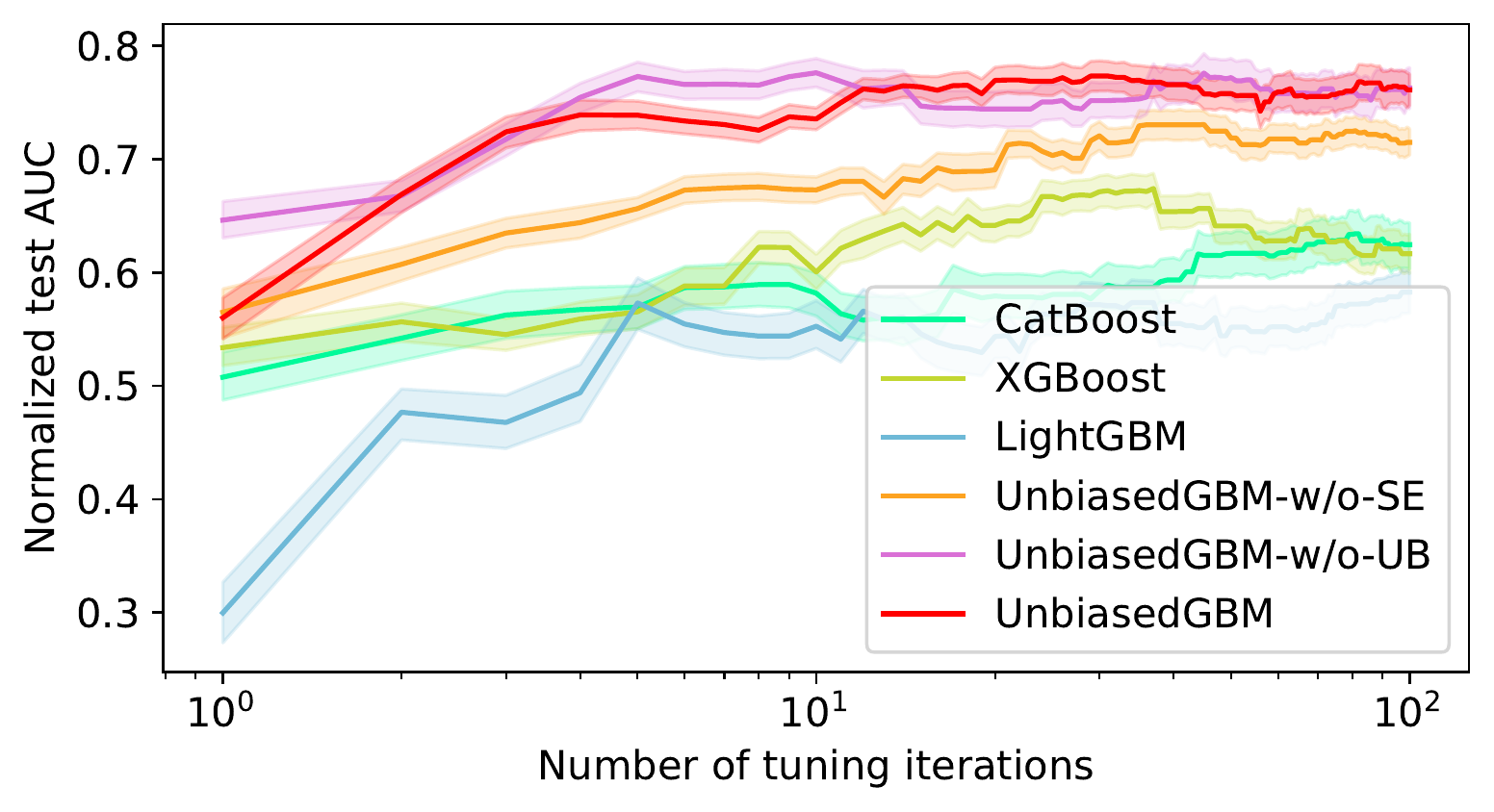}
        \end{minipage}%
    }%
    
    \subfigure[10 medium-scale datasets with only numerical features.]{
        \begin{minipage}[t]{0.48\linewidth}
        \centering
        \includegraphics[width=\textwidth]{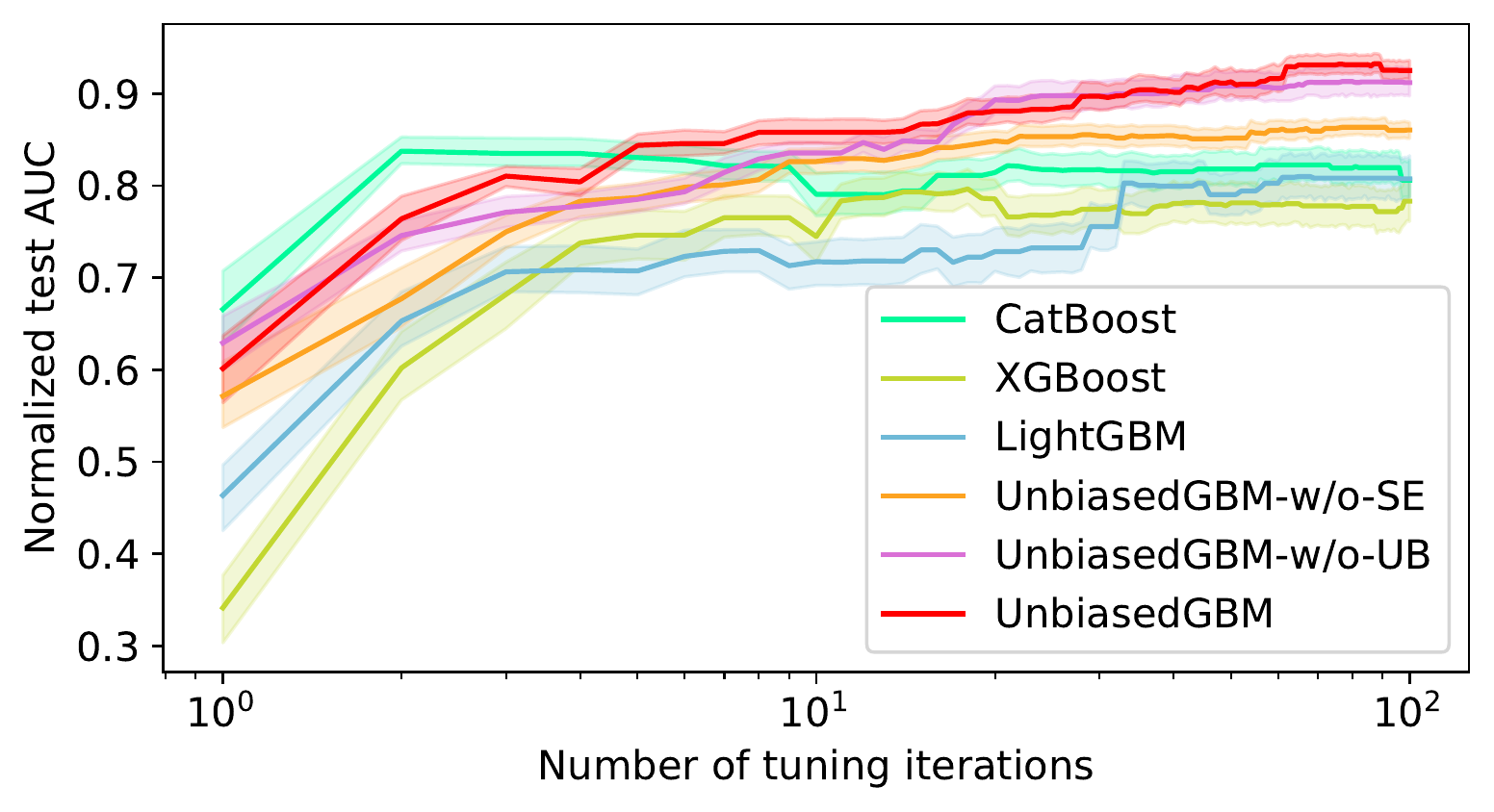}
        \end{minipage}
    }%
    \subfigure[10 medium-scale datasets with num. and cat. features.]{
        \begin{minipage}[t]{0.48\linewidth}
        \centering
        \includegraphics[width=\textwidth]{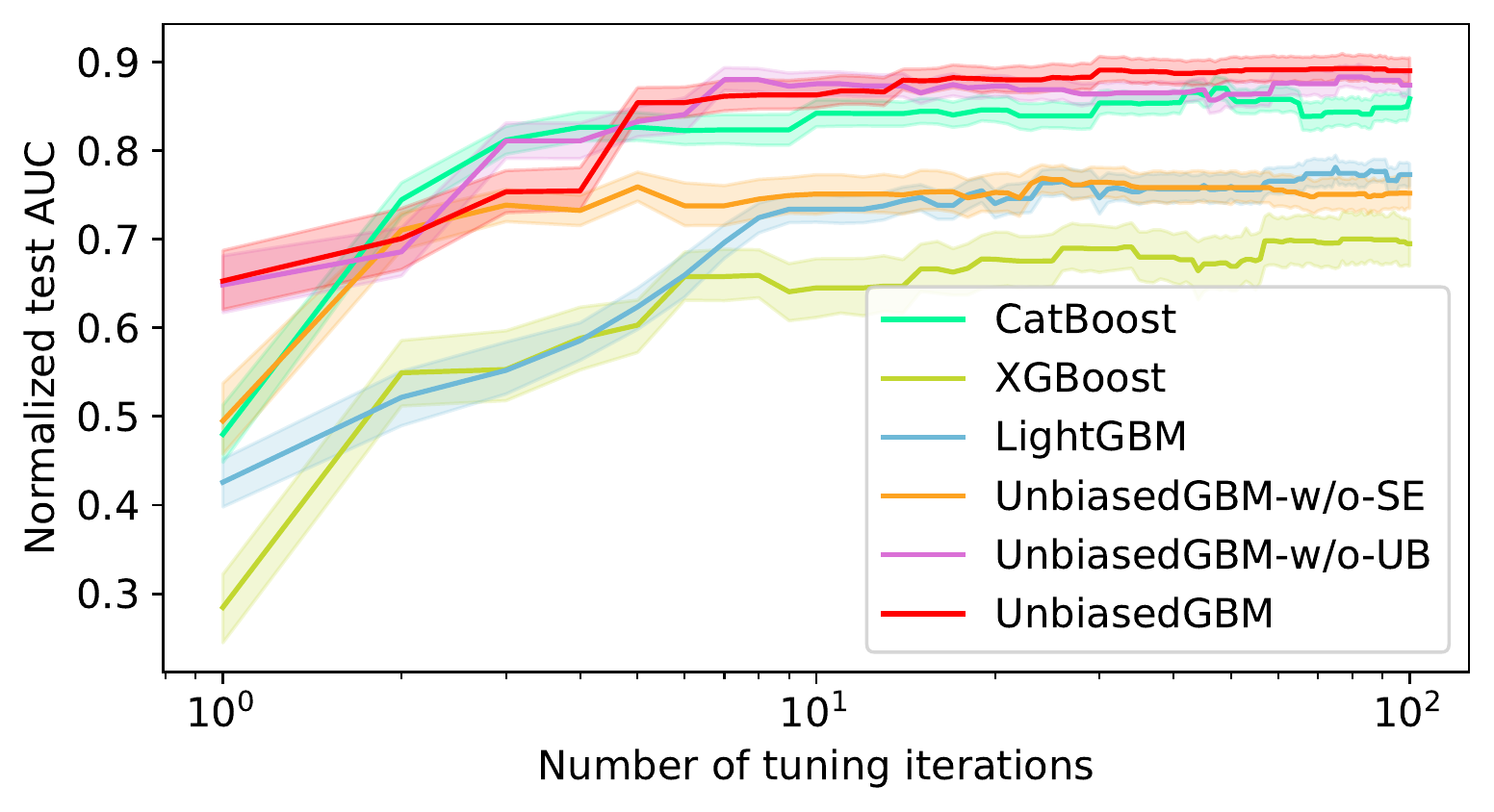}
        \end{minipage}
    }%

\caption{Comparison of UnbiasedGBM with other baseline methods on four types of datasets. Each value corresponds the normalized test AUC of the best model (on the validation set) after a specific number of tuning iterations, averaged on all the datasets. The shaded area presents the variance of the scores.}
\label{fig: comparison}
\end{figure*}

In this section, we aim at answering two questions through extensive experiments:
\begin{itemize}
    \item \textbf{Q1.} How does UnbiasedGBM perform compared with well-developed GBDT implementations such as XGBoost, LightGBM, and CatBoost?
    \item \textbf{Q2.} How does the proposed unbiased gain perform in terms of feature selection compared with existing feature importance methods?
\end{itemize}

\subsection{Datasets} \label{subsec: datasets}
We collect 60 classification datasets in various application domains provided by Kaggle, UCI~\cite{UCI}, and OpenML~\cite{DBLP:journals/sigkdd/OpenML} platforms. We select datasets according to the following criteria: \textbf{1) Real-world data}. We remove artificial datasets that are designed to test specific models. \textbf{2) Not high dimensional}. We remove datasets with $m/n$ ratio above $1$. \textbf{3) 
 Not too small}. We remove datasets with too few samples ($<500$). \textbf{4) Not too easy}. We remove datasets if a LightGBM with the default hyperparameters can reach a score larger than $0.95$. The detailed properties of datasets are presented in Appendix~\ref{app: sec: dataset}.

\subsection{Q1. UnbiasedGBM}
In this subsection, we answer the Q1 question by comparing UnbiasedGBM with XGBoost, LightGBM, and CatBoost using extensive experiments.

\subsubsection{Evaluation Metrics}

\label{sec:normAUC}

We use the area under the ROC curve (AUC) in the test set to measure the model performance. In order to aggregate results across datasets of different difficulty, we employ a metric similar to the distance to the minimum, which is introduced in~\cite{wistuba2015learning} and used in~\cite{feurer2020auto,why-tree}. This metric normalize each test AUC between 0 and 1 via a min-max normalization using the worst AUC and the best AUC of all the models on the dataset.

\begin{figure*}[ht]
\centering
    \subfigure[UnbiasedGBM vs. LightGBM]{
        \begin{minipage}[t]{0.3\linewidth}
        \centering
        \includegraphics[width=\textwidth]{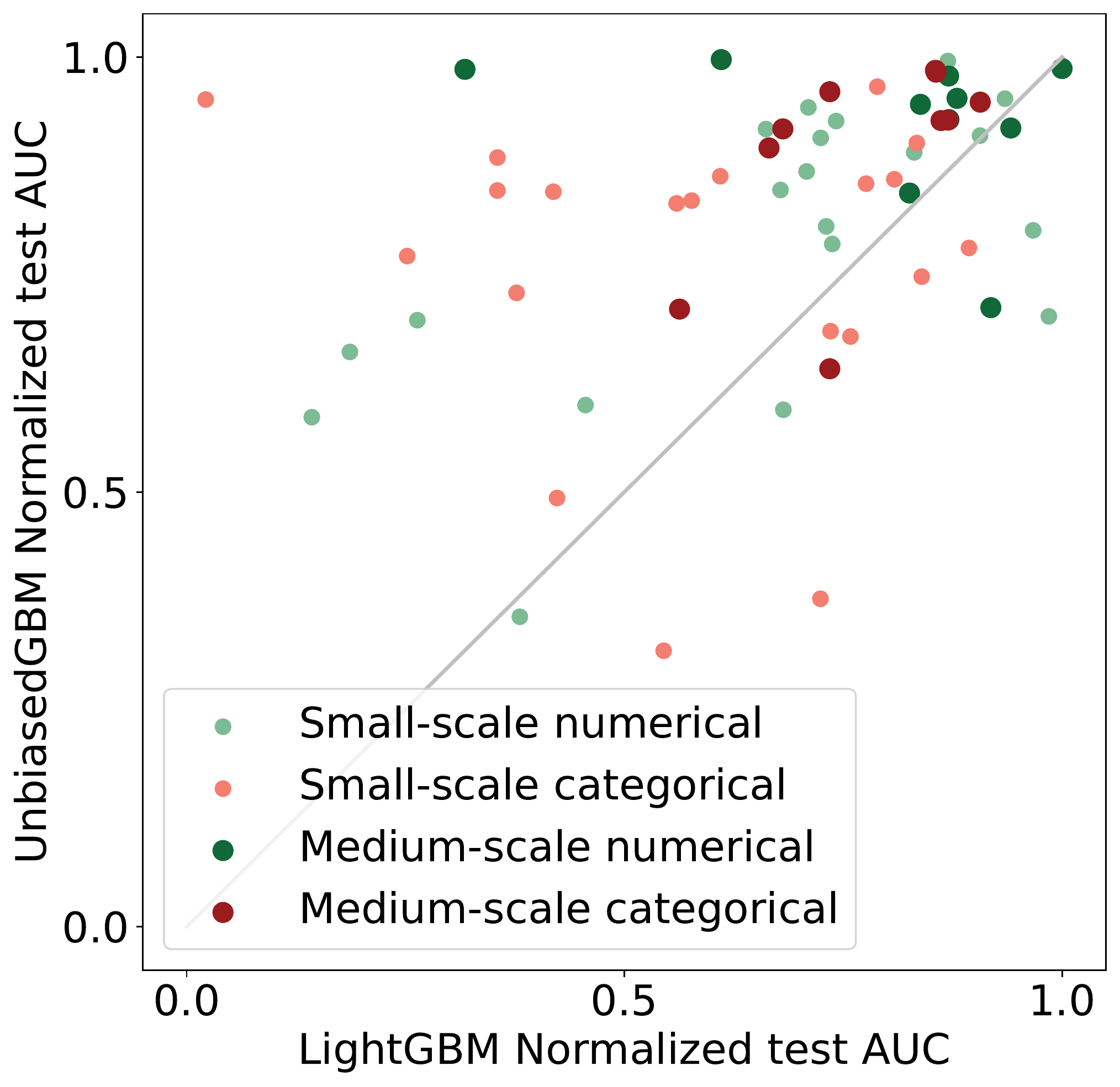}
        \end{minipage}%
    } $\quad$%
    \subfigure[UnbiasedGBM vs. XGBoost]{
        \begin{minipage}[t]{0.3\linewidth}
        \centering
        \includegraphics[width=\textwidth]{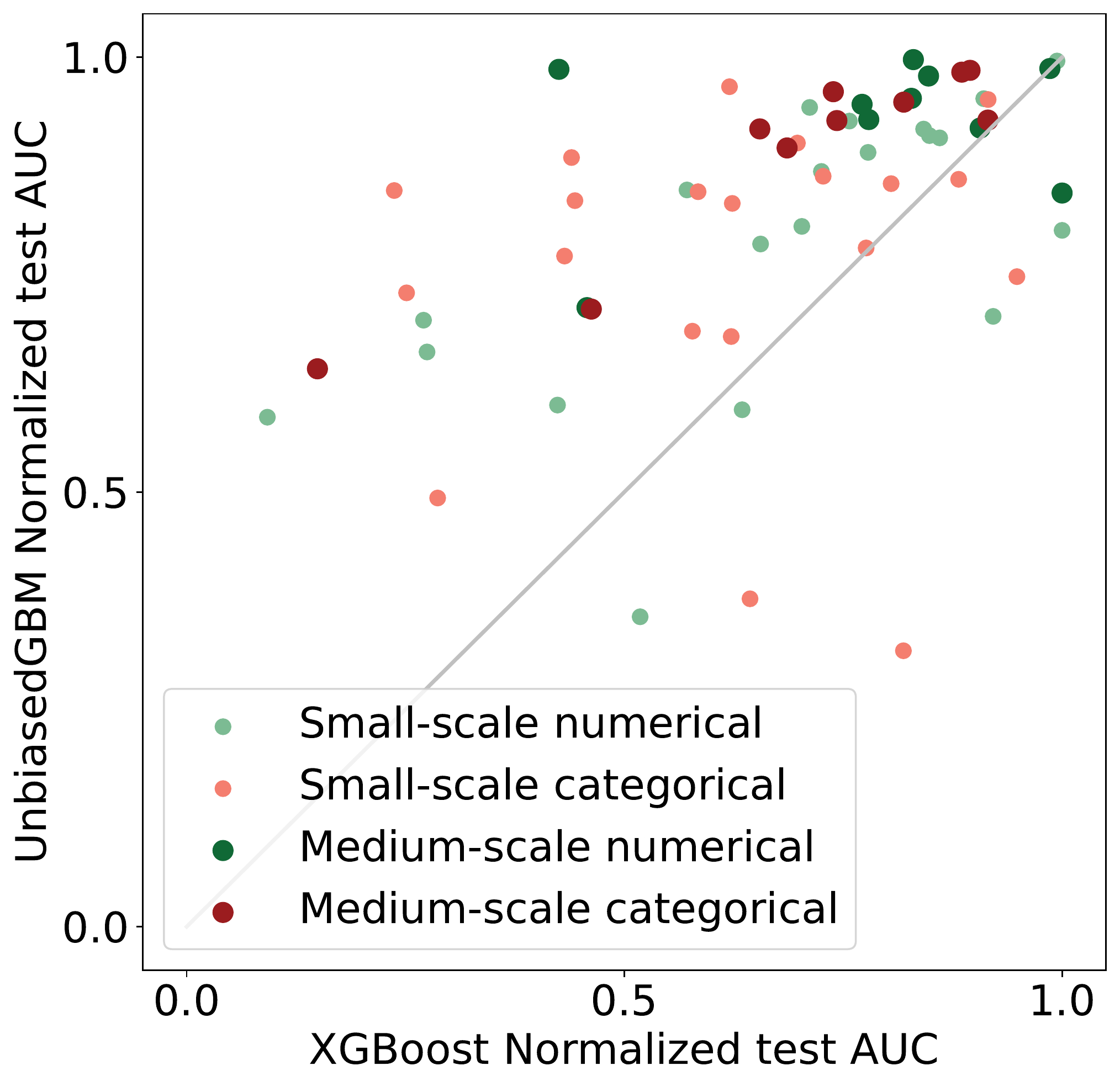}
        \end{minipage}%
    } $\quad$%
    \subfigure[UnbiasedGBM vs. CatBoost]{
        \begin{minipage}[t]{0.3\linewidth}
        \centering
        \includegraphics[width=\textwidth]{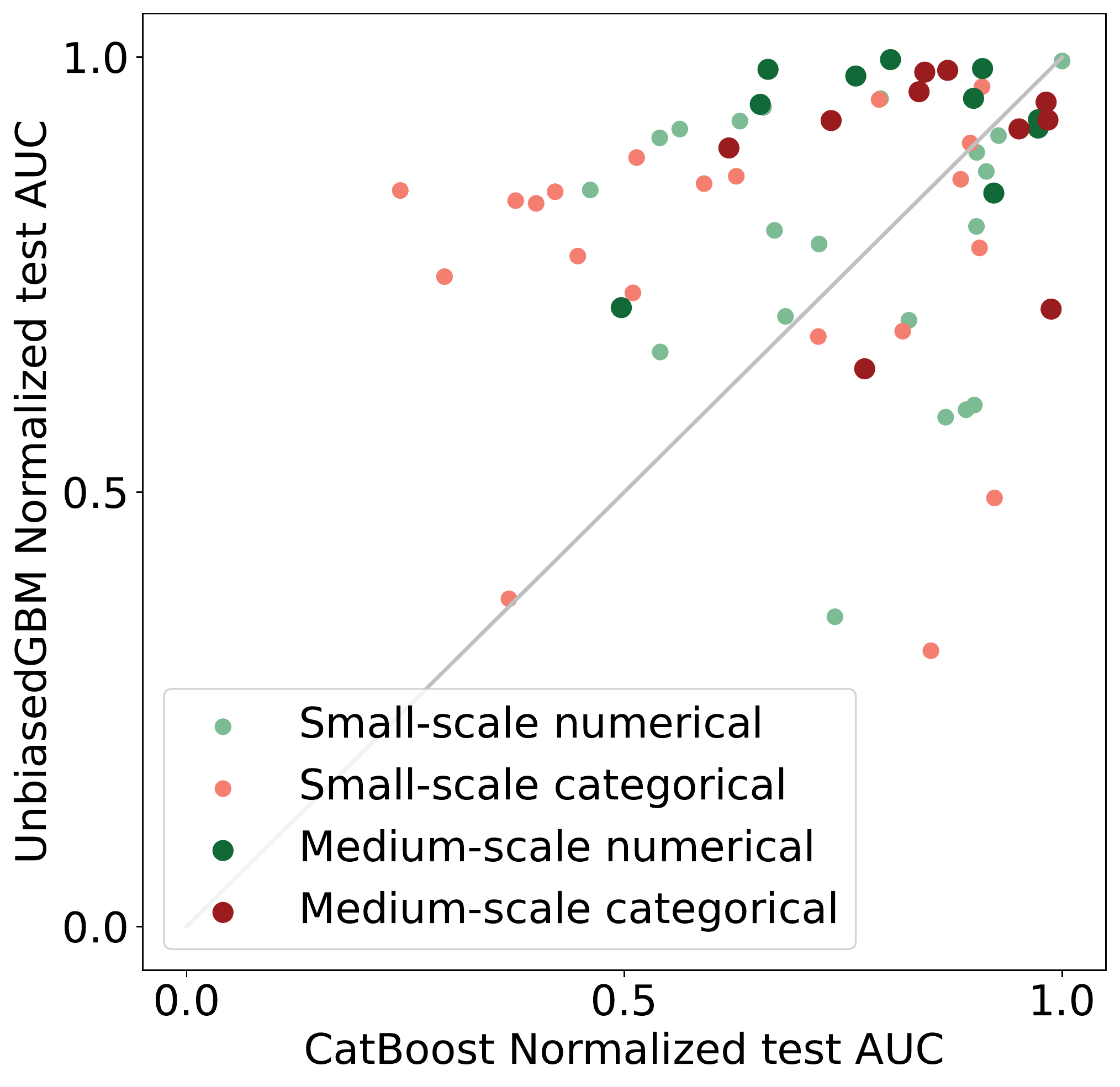}
        \end{minipage}%
    } $\quad$%

\caption{Comparison of UnbiasedGBM with LightGBM, XGBoost and CatBoost. Each dot denotes a dataset. The normalized test AUC is higher the better. ``numerical'' means the dataset only contains numerical features. ``categorical'' means the dataset contains both numerical and categorical features.}
\label{fig: dot-comparison}
\end{figure*}

\begin{figure}[ht]
    \centering
    \includegraphics[width=0.48\textwidth]{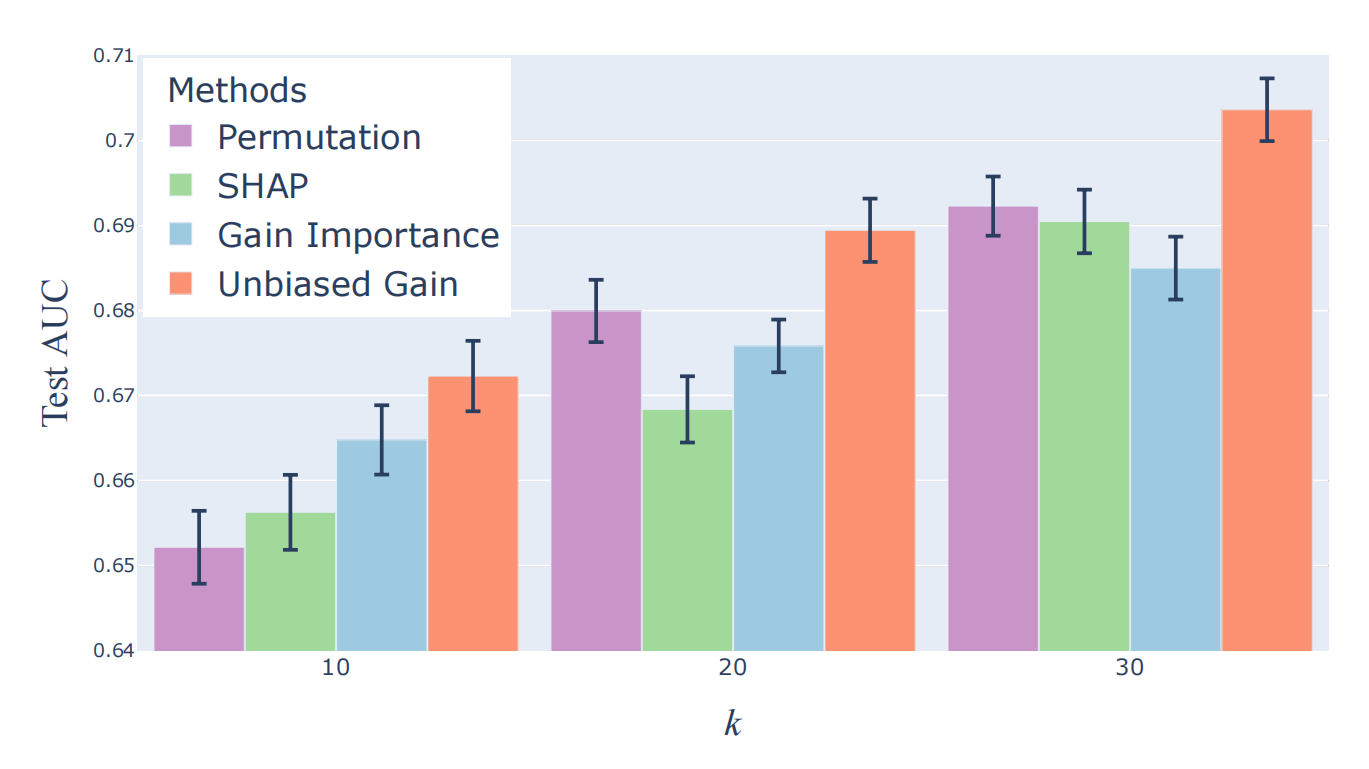}
    \caption{Comparison of different feature importance methods in feature selection. We report the AUC on the test set of the model using top $k\%$ selected features according to the feature importance.}
    \label{fig: feature selection}
\end{figure}

\subsubsection{Baseline Methods}
We compare with the following baseline methods:
\begin{itemize}
    \item \textbf{XGBoost}~\cite{xgboost}.
    \item \textbf{LightGBM}~\cite{lightgbm}. 
    \item \textbf{CatBoost}~\cite{catboost}. 
    \item \textbf{UnbiasedGBM-w/o-SE}. UnbiasedGBM without separating the determination of the best split of each feature and the feature to split. One of two validation sets is merged with the subtraining set.
    \item \textbf{UnbiasedGBM-w/o-UB}. UnbiasedGBM without computing the unbiased gain $\widetilde{\mathrm{Gain}}_{ub}$ to measure the generalization performance of the best split.  Two validation sets are merged into one to determine the best feature to split and perform early stopping.
\end{itemize}
For each method, we perform hyperparameter optimization using the popular Optuna~\cite{optuna} Python package. See more details in Appendix~\ref{app: sec: hpo}.

\begin{table}[]
    \centering
    \resizebox{\linewidth}{!}{
    \begin{tabular}{ccccc} \toprule
         & XGBoost & LightGBM & CatBoost & UnbiasedGBM \\ \midrule
         Average Rank & 3.00 & 2.85 & 2.43 & 1.72 \\ 
         p-value & $\le 10^{-3}$ & $\le 10^{-3}$ & 0.013 & - \\ \bottomrule
    \end{tabular}
    }
    \caption{The average rank (where lower is better) over 60 datasets and the p-value of Nemenyi test between UnbiasedGBM and the baseline methods.}
    \label{tab: average rank}
    \vspace{-0.2cm}
\end{table}

\subsubsection{Results}
In order to better present the advantage of UnbiasedGBM on datasets with different properties, we classify 60 datasets into four types of datasets, including small-scale (datasets with less than 4000 samples) or medium-scale datasets with only numerical features or both numerical and categorical features. We present the results in Figure \ref{fig: comparison}. The x-axis is the number of tuning iterations, visualizing the influence of the tuning budget on the model performance. We can see that UnbiasedGBM significantly outperforms XGBoost, LightGBM, and CatBoost in both small and medium datasets. In addition, UnbiasedGBM is effective even if there is only numerical features in the datasets. Categorical features are not the only source of performance improvement in UnbiasedGBM. We also visualize the comparison of each dataset in Figure \ref{fig: dot-comparison} that demonstrates the improvement of our method. We leverage the Nemenyi test~\cite{DBLP:journals/jmlr/Demsar06} to perform statistical analyses using the rank of each method after hyper-parameter tuning of 100 iterations on 60 datasets. We present the results in Table \ref{tab: average rank}, where the Nemenyi test p-values show that UnbiasedGBM significantly outperforms the baselines. Moreover, comparisons between UnbiasedGBM, UnbiasedGBM-w/o-SE, and UnbiasedGBM-w/o-UB demonstrate that separating the determination of the best split of each feature and the feature to split is the primary source of improvement in UnbiasedGBM. In most cases, computing the unbiased gain to evaluate generalization performance of the best split can also result in performance improvement.

\subsection{Q2. Unbiased Gain}
In this subsection, we demonstrate the performance of unbiased gain in feature selection.
\subsubsection{Baseline Methods}
We compare unbiased gain with different feature importance measurements:
\begin{itemize}
    \item Gain importance~\cite{breiman2017classification}. 
    \item Permutation feature importance (PFI)~\cite{breiman2001random}. 
    \item SHAP~\cite{TreeSHAP}. 
\end{itemize}
\subsubsection{Evaluation}
We follow the standard approach~\cite{forman2003extensive} to evaluate  different feature importance measurements in feature selection. For a given dataset, we first estimate the feature importance on the training set. Then, we select top $k\%$ features according to the feature importance, where $k\in \{10,20,30\}$. Next, we build a GBDT model according to the selected feature subset. Finally, we calculate the AUC of the model on the test set. Higher AUC indicates that the feature importance performs better in feature selection.
\subsubsection{Results}
We evaluate these methods on 14 out of 60 datasets with more than 30 features. Datasets with too few features may not need feature selection. We consider selecting top $k\%$ features for $k\in\{10,20,30\}$. For each method, we report the mean and variance of the test AUC across these 14 datasets. The results are presented in Figure \ref{fig: feature selection}. We can see that unbiased gain achieves better average performance than baseline methods in feature selection.

\begin{table}[]
    \centering
    \resizebox{\linewidth}{!}{
    \begin{tabular}{ccc} \toprule
          & LightGBM & UnbiasedGBM \\ \midrule
         Full feature set & $0.779\scriptscriptstyle \pm \scriptstyle 0.000$ & $0.809 \scriptscriptstyle \pm \scriptstyle 0.003$ \\
         Remove ``nHM'' with 11 categories & $0.772\scriptscriptstyle \pm \scriptstyle 0.000$ & $0.797 \scriptscriptstyle \pm \scriptstyle 0.003$ \\ 
         Remove ``PCD'' with 224 categories & $0.787\scriptscriptstyle \pm \scriptstyle 0.000$ & $0.811 \scriptscriptstyle \pm \scriptstyle 0.002$ \\ \bottomrule 
    \end{tabular}
    }
    \caption{An example of the QSAR Bioconcentration dataset. LightGBM overfits on the ``PCD'' feature with many categories, because removing the feature brings significant improvement in the test AUC. UnbiasedGBM addresses the overfitting issue, because it has better test AUC than LightGBM when using the full feature set, and removing the ``PCD'' feature brings an insignificant difference.}
    \label{tab:performance}
\end{table}

\subsection{Analyses of Features with Many Categories}

We present an analysis of the QSAR Bioconcentration dataset in Table \ref{tab:performance} to show that UnbiasedGBM can address the overfitting issue of LightGBM on categorical features with many categories. The details are in the caption of Table \ref{tab:performance}.

\section{Conclusion}
In this paper, we investigate the bias in GBDT and the consequent interpretability and overfitting issues. We give a fine-grained analysis of bias in GBDT. Based on the analysis, we propose the unbiased gain and UnbiasedGBM to address the interpretability and overfitting issues. Extensive experiments on 60 datasets show that UnbiasedGBM has better average performance than XGBoost, LightGBM, and Catboost and unbiased gain can outperform popular feature importance estimation methods in feature selection.
\newpage

\section*{Acknowledgements}
The authors are supported in part by the National Natural Science Foundation of China Grant 62161146004, Turing AI Institute of Nanjing and Xi'an Institute for Interdisciplinary Information Core Technology.

\section*{Contribution Statement}
This paper is the result of collaborative work between Zheyu Zhang and Tianping Zhang, who contributed equally to the conception, implementation, experimentation, and paper writing. Jian Li served as the corresponding author, contributing to the overall idea of the project as well as providing computing resources.

\bibliographystyle{named}
\bibliography{ijcai23}

\newpage

\begin{linenomath}
$$\quad$$
\end{linenomath}

\newpage

\appendix

\section{Proof and Discussion for Theorem 1} \label{app: sec: proof 1}


Theorem~\ref{thm: gain ge zero} reveals that the split gain on node $I$ with split $\theta=(j,s)$, written in
\begin{linenomath}
$$\widetilde {\mathrm{Gain}}(I,\theta) = \widetilde {\mathcal L}(I) - \widetilde {\mathcal L}(I_L) - \widetilde {\mathcal L}(I_R)$$
\end{linenomath}
is always non-negative, where
\begin{linenomath}
$$\widetilde {\mathcal L}(I) = -\frac{1}{2} \frac{\left( \frac{1}{n_I} \sum_{i\in I} g_i\right)^2}{\frac{1}{n_I} \sum_{i\in I} h_i} \frac{n_I}{n} = -\frac{1}{2n} \frac{G_I^2}{H_I}.$$
\end{linenomath}

\noindent \textbf{Theorem 1.}
For a dataset $(X, Y)$ sampled from a distribution $\mathcal{T}$, for any split $\theta$ of node $I$ on a given feature $X_j$, we always have
\begin{linenomath}
$$ \widetilde {\mathrm{Gain}}(I,\theta)\geq 0.$$
\end{linenomath}

\begin{proof}
First, rewrite $\widetilde {\mathcal L}(I)$ with the optimization setting:
\begin{linenomath}
\begin{align*}
    \widetilde {\mathcal L}(I) & = - \frac{1}{2n}\frac{G_I^2}{H_I} \\
    & = \frac{1}{n} \min_w \left(\frac{1}{2}H_I w^2 
+ G_I w\right)\\
    & = \frac{1}{n} \min_w \sum_{i\in I}\left(\frac{1}{2}h_i w^2  + g_i w\right).
\end{align*}
\end{linenomath}
Since $I=I_L\cup I_R$, the total of the optimal loss of $I_L$ and $I_R$ is smaller than the optimal loss of $I$:
\begin{linenomath}
\begin{equation}
\begin{aligned}
    \widetilde {\mathrm{Gain}}(I, \theta) = &\ \widetilde {\mathcal L}(I) - \widetilde {\mathcal L}(I_L) - \widetilde {\mathcal L}(I_R) \\
    = & \frac{1}{n} \Big(\min_{w} \sum_{i \in I} l_i(w)\\
      & - \min_{w_L} \sum_{i \in I_L} l_i(w_L) - \min_{w_R} \sum_{i \in I_R} l_i(w_R) \Big) \label{eq: proof1}\\
    \ge &\ 0,
\end{aligned}
\end{equation}
\end{linenomath}
where $l_i(w) = \frac{1}{2} h_i w^2 + g_i w$.
\end{proof}

\paragraph{\textbf{Discussion.}} Theorem~\ref{thm: gain ge zero} shows that $\widetilde {\mathrm{Gain}}(I, \theta)$ is always non-negative. From Eq~\ref{eq: proof1}, we know that $\widetilde {\mathrm{Gain}}(I, \theta)=0$ if and only if $w^*=w_L^*=w_R^*$, which is equivalent to
\begin{linenomath}
$$\frac{G_L}{H_L} = \frac{G_R}{H_R}.$$
\end{linenomath}
This is a sufficient and necessary condition for $\widetilde {\mathrm{Gain}}(I, \theta)=0$, which is a very rare case in the applications.




\section{Proof and Explanation for Theorem 2}
\label{app: sec: proof}

Assume we have a training dataset $\mathcal D=\{(\mathbf x_i,y_i)\}$ and a validation dataset $\mathcal D'=\{(\mathbf x_i',y_i')\}$. For a given leaf node $I$ and a given split $I=I_L\cup I_R$, there are $n_I$, $n_L$, $n_R$ training examples and $n_I'$, $n_L'$, $n_R'$ validation examples that fall into leaf nodes $I$, $I_L$, and $I_R$. First, we estimate $\mu_g(I)$ using the training examples
\begin{linenomath}
$$
\hat{\mu}_g(I)=\frac{1}{n_I}G_I=\frac{1}{n_I}\sum_{q(x_i)=I}g_i.
$$
\end{linenomath}

Then, we randomly select $k$ examples from $n_I'$ validation examples, where $k=\min(n_L', n_R')$. Next, we estimate $\mu_g(I)$ and $\mu_h(I)$ using $k$ randomly selected validation examples
\begin{linenomath}
\begin{align*}
    \hat{\mu}_g'(I)=& \frac{1}{k}G_I'=\frac{1}{k}\sum_{q(x_i')=I}g_i'\cdot \delta(I,i), \\
    \hat{\mu}_h'(I)=& \frac{1}{k}H_I'=\frac{1}{k}\sum_{q(x_i')=I}h_i'\cdot \delta(I,i),
\end{align*}
\end{linenomath}
where $\delta(I,i)$ is a binary indicator showing whether a validation sample has been selected. Finally we can calculate the loss of leaf node $I$ by
\begin{linenomath}
$$
\widetilde{\mathcal{L}}(I)=\frac{1}{2}\frac{\hat{\mu}_g(I)\cdot \hat{\mu}_g'(I)}{\hat{\mu}_h'(I)}\cdot \frac{n_I}{n}=-\frac{1}{2n}G_I\cdot \frac{G_I'}{H_I'}\ .
$$
\end{linenomath}
Here, $G_I$ is computed using the training set while $G_I'$ and $H_I'$ are computed using the validation set. We can also calculate $\widetilde{\mathcal{L}}(I_L)$ and $\widetilde{\mathcal{L}}(I_R)$ in a similar way (the number of selected validation example $k$ is the same for $I$, $I_L$, and $I_R$). Finally, the unbiased gain is calculated as
\begin{linenomath}
\begin{equation}
\label{eqn: unbiased gain}
    \widetilde{\mathrm{Gain}}_{\mathrm{ub}}(I,\theta) = 
      \widetilde{\mathcal{L}}(I) -
     \widetilde{\mathcal{L}}(I_L) - \widetilde{\mathcal{L}}(I_R).
\end{equation}
\end{linenomath}

\noindent \textbf{Theorem 2.}
For a feature $X_j$, a leaf node $I$, and \zzy{a split $\theta$}, if $X_j$ is marginally independent of $y$ within the region defined by the leaf node $I$, then
\begin{linenomath}
$$\mathbb E_{\mathcal D'} \left[ \widetilde{\mathrm{Gain}}_{\mathrm{ub}}(I,\theta) \right] = 0.$$
\end{linenomath}

\begin{proof}
Since $\hat \mu_g'(I)$, $\hat \mu_g'(I_L)$, $\hat \mu_g'(I_R)$ and $\hat{\mu}_h'(I)$, $\hat{\mu}_h'(I_L)$, $\hat{\mu}_h'(I_R)$ are all estimated by the same number of $k$ samples, we have
\begin{linenomath}
\begin{equation}
\label{app: eqn: proof_unbiased_gain}
\forall k, \mathbb E \left[ \frac{\hat \mu_g'(I)}{\hat{\mu}_h'(I)} \Bigg| k\right] = \mathbb E \left[ \frac{\hat \mu_g'(I_L)}{\hat{\mu}_h'(I_L)} \Bigg| k\right] = \mathbb E \left[ \frac{\hat \mu_g'(I_R)}{\hat{\mu}_h'(I_R)} \Bigg| k\right],    
\end{equation}
\end{linenomath}

where $\mathbb E$ is short for $\mathbb E_{\mathcal D'}$. Hence
\begin{linenomath}
\begin{align*}
    \mathbb E \left[ \widetilde{\mathrm{Gain}}_{\mathrm{ub}} \right] = &\ 
\mathbb E \left[ \widetilde{\mathcal{L}}_{\mathrm{ub}}(I)\right] -
\mathbb E \left[ \widetilde{\mathcal{L}}_{\mathrm{ub}}(I_L)\right] - 
\mathbb E \left[ \widetilde{\mathcal{L}}_{\mathrm{ub}}(I_R)\right] \\
=
&\ - \frac{G_I}{2n} \mathbb E \left[ \frac{\hat \mu_g'(I)}{\hat{\mu}_h'(I)} \right] \\
&\ + \frac{G_L}{2n} \mathbb E \left[ \frac{\hat \mu_g'(I_L)}{\hat{\mu}_h'(I_L)} \right]
+ \frac{G_R}{2n} \mathbb E \left[ \frac{\hat \mu_g'(I_R)}{\hat{\mu}_h'(I_R)} \right] \\
=
&\ \frac{G_L+G_R-G_I}{2n} \sum_k P(k) \mathbb E \left[ \frac{\hat \mu_g'(I)}{\hat{\mu}_h'(I)} \Bigg| k \right] \\
= &\ 0,
\end{align*}
\end{linenomath}
\end{proof}


\subsection{The Motivation Behind the Unbiased Gain}
\paragraph{\textbf{Why do we need an additional validation set?}} The intuitive rationale behind this is that we should not find the optimal split and evaluate the optimal split using the same set of data.

\paragraph{\textbf{Can we re-calculate the reduction in loss using the validation set?}} An intuitive way of using the validation set is to fix the tree structure and re-calculate the reduction in loss using the validation set. However, for a split on an uninformative feature, the split gain evaluated using the validation set is expected to be negative (instead of zero)~\cite{unbiased}. 

\paragraph{Why do we need to randomly select $k$ samples when calculating the unbiased gain?} The Eq. \ref{app: eqn: proof_unbiased_gain} does not hold if $\hat \mu_g'(I)$, $\hat \mu_g'(I_L)$, $\hat \mu_g'(I_R)$ and $\hat{\mu}_h'(I)$, $\hat{\mu}_h'(I_L)$, $\hat{\mu}_h'(I_R)$ are estimated by different number of samples, and thus we cannot derive the unbiased property of the gain estimation.




\section{Datasets}
\label{app: sec: dataset}

The 60 datasets are collected from a repository of 654 datasets from Kaggle, UCI, and OpenML platforms. In order to collect datasets of different types (datasets with different scales and whether the dataset has categorical features), we select the datasets according to Algorithm \ref{alg:datasets}. Table~\ref{tab:cla_large_cat}~\ref{tab:cla_large_num}~\ref{tab:cla_small_cat}~\ref{tab:cla_small_num} show the datasets we collected and used in our experiment.

\begin{algorithm}[t]
    \caption{Select 60 Datasets}
    \label{alg:datasets}
    \textbf{Input}: A Reservoir of 654 Datasets \\
    \textbf{Output}: Selected 60 datasets.
    
    \begin{algorithmic}[1] 
\STATE Result $\leftarrow \emptyset$
\FOR{$\mathcal D$ in sorted(Reservoir, ascent order by instance)}
    \STATE Type $\leftarrow \left(\mathrm{Scale}(\mathcal D), \mathrm{Categorical}(\mathcal D)\right)$
    \IF{datasets of this Type is enough}
        \STATE \textbf{Continue}
    \ENDIF
    \STATE Model $\leftarrow$ Baseline($\mathcal D$)
    \STATE Pred $\leftarrow$ Model.predict($\mathcal D'$)
    \STATE Score $\leftarrow$ AUC($\mathcal D'$, Pred)
    \IF{Score $> 0.95$ \textbf{or} Score $< 0.55$}
        \STATE \textbf{Continue}
    \ENDIF
    \STATE Result $\leftarrow$ Result $\cup \{\mathcal D\}$
\ENDFOR
\RETURN Result
    \end{algorithmic}
\end{algorithm}

\begin{table*} \centering \begin{tabular}{llrrr} \toprule
 source & name & sample & num feat & cat feat \\ \midrule 
kaggle & Churn Modelling & 10000 & 8 & 3 \\ 
kaggle & Online Shopper's Intention & 12330 & 14 & 3 \\ 
kaggle & HR analysis & 18359 & 3 & 10 \\ 
kaggle & Donors-Prediction & 19372 & 42 & 6 \\ 
kaggle & aam avaliacao dataset & 25697 & 12 & 10 \\ 
openml & airlines & 26969 & 4 & 3 \\ 
kaggle & Income Predictions Dataset(2 class classification) & 30162 & 6 & 8 \\ 
kaggle & Success of Bank Telemarketing Data & 30477 & 1 & 6 \\ 
kaggle & Term Deposit Prediction Data Set & 31647 & 7 & 9 \\ 
UCI & Adult & 32561 & 6 & 8 \\ 
\bottomrule \end{tabular} \caption{Medium-scale categorical datasets.} \label{tab:cla_large_cat} \end{table*}

\begin{table*} \centering \begin{tabular}{llrrr} \toprule
 source & name & sample & num feat & cat feat \\ \midrule 
kaggle & Amsterdam - AirBnb & 10498 & 16 & 0 \\ 
UCI & Firm-Teacher Clave-Direction Classification & 10799 & 16 & 0 \\ 
openml & jm1 & 10885 & 21 & 0 \\ 
openml & eye movements & 10936 & 26 & 0 \\ 
kaggle & Kyivstar Big Data test & 11583 & 45 & 0 \\ 
kaggle & Flower Type Prediction Machine Hack & 12666 & 6 & 0 \\ 
openml & test dataset & 15547 & 60 & 0 \\ 
openml & elevators & 16599 & 18 & 0 \\ 
openml & MagicTelescope & 19020 & 10 & 0 \\ 
kaggle & Web Club Recruitment 2018 & 20000 & 23 & 0 \\ 
\bottomrule \end{tabular} \caption{Medium-scale numerical datasets.} \label{tab:cla_large_num} \end{table*}

\begin{table*} \centering \begin{tabular}{llrrr} \toprule
 source & name & sample & num feat & cat feat \\ \midrule 
UCI & ILPD (Indian Liver Patient Dataset) & 583 & 9 & 1 \\ 
kaggle & Credit Card Approval & 590 & 6 & 9 \\ 
kaggle & Analytics Vidhya Loan Prediction & 614 & 5 & 6 \\ 
kaggle & Student Alcohol Consumption & 649 & 13 & 17 \\ 
UCI & QSAR Bioconcentration classes dataset & 779 & 10 & 1 \\ 
kaggle & The Estonia Disaster Passenger List & 989 & 1 & 5 \\ 
UCI & Statlog (German Credit Data) & 999 & 7 & 13 \\ 
openml & credit-g & 1000 & 7 & 13 \\ 
kaggle & Employee Attrition & 1029 & 24 & 7 \\ 
kaggle & Train Crowd Density & 1284 & 7 & 9 \\ 
UCI & Yeast & 1484 & 8 & 1 \\ 
UCI & Drug consumption (quantified) & 1885 & 13 & 18 \\ 
kaggle & RMS Lusitania Complete Passenger Manifest & 1961 & 1 & 11 \\ 
kaggle & Marketing Campaign & 2240 & 23 & 3 \\ 
UCI & seismic-bumps & 2584 & 11 & 4 \\ 
kaggle & Telecom Churn Dataset & 2666 & 16 & 3 \\ 
kaggle & Well log facies dataset & 3232 & 8 & 2 \\ 
kaggle & Client churn rate in Telecom sector & 3333 & 16 & 3 \\ 
kaggle & Cardiovascular Study Dataset & 3390 & 13 & 2 \\ 
kaggle & Campus France Rouen 2019 admission & 3585 & 2 & 6 \\ 
\bottomrule \end{tabular} \caption{Small-scale categorical datasets.} \label{tab:cla_small_cat} \end{table*}

\begin{table*} \centering \begin{tabular}{llrrr} \toprule
 source & name & sample & num feat & cat feat \\ \midrule 
openml & fri c3 1000 10 & 1000 & 10 & 0 \\ 
kaggle & Customer Classification & 1000 & 11 & 0 \\ 
openml & autoUniv-au1-1000 & 1000 & 20 & 0 \\ 
openml & fri c0 1000 50 & 1000 & 50 & 0 \\ 
openml & fri c1 1000 50 & 1000 & 50 & 0 \\ 
openml & rmftsa sleepdata & 1024 & 2 & 0 \\ 
openml & PizzaCutter3 & 1043 & 37 & 0 \\ 
UCI & QSAR biodegradation & 1055 & 41 & 0 \\ 
openml & PieChart3 & 1077 & 37 & 0 \\ 
kaggle & Credit Risk Classification Dataset & 1125 & 11 & 0 \\ 
UCI & Diabetic Retinopathy Debrecen Data Set & 1151 & 19 & 0 \\ 
kaggle & Heart Disease Dataset (Comprehensive) & 1190 & 11 & 0 \\ 
openml & pc4 & 1458 & 37 & 0 \\ 
kaggle & HR-attrition-EDA & 1470 & 44 & 0 \\ 
UCI & Contraceptive Method Choice & 1473 & 9 & 0 \\ 
kaggle & Bangla Music Dataset & 1742 & 29 & 0 \\ 
kaggle & Diabetes Data Set & 2000 & 8 & 0 \\ 
openml & kc1 & 2109 & 21 & 0 \\ 
openml & Titanic & 2201 & 3 & 0 \\ 
openml & space ga & 3107 & 6 & 0 \\ 
\bottomrule \end{tabular} \caption{Small-scale numerical datasets.} \label{tab:cla_small_num} \end{table*}

\section{Hyperparameter Optimization}
\label{app: sec: hpo}

Table~\ref{tab:hyper} shows the hyperparameter spaces for each method. Optuna tunes all methods over 100 epochs.

\begin{table*} \centering \begin{tabular}{lccc} \toprule
    method & Hyperparameter & range & log \\ \midrule 
    XGBoost & n\_estimators & 200$\sim$3000(small)/6000(medium) & True \\
            & learning\_rate & 0.005$\sim$0.05 & True \\
            & min\_child\_weight & 2$\sim$20 & True \\
            & gamma & 0$\sim$0.1 & False \\ \midrule
    LightGBM & n\_estimators & 200$\sim$3000(small)/6000(medium) & True \\
            & learning\_rate & 0.005$\sim$0.05 & True \\
            & min\_child\_weight & 2$\sim$20 & True \\
            & min\_split\_gain & 0$\sim$0.1 & False \\ \midrule
    CatBoost & n\_estimators & 200$\sim$3000(small)/6000(medium) & True \\
            & learning\_rate & 0.005$\sim$0.05 & True \\
            & min\_data\_in\_leaf & 2$\sim$20 & True \\
            & l2\_leaf\_reg & 0$\sim$0.1 & False \\ \midrule
    UnbiasedGBM & n\_estimators & 200$\sim$3000(small)/6000(medium) & True \\
            & learning\_rate & 0.005$\sim$0.05 & True \\
            & min\_data\_in\_leaf & 2$\sim$20 & True \\
            & min\_split\_gain & -0.1$\sim$0.1 & False \\
\bottomrule \end{tabular} \caption{Hyperparameters.} \label{tab:hyper} \end{table*}

\section{Implementation Details}

\label{app: sec: ImpDetail}

\subsection{Split Finding Algorithm}

The idea of UnbiasedGBM can be incorporated in existing split finding algorithms. The current implementation of UnbiasedGBM is based on XGBoost. Algorithm~\ref{alg:algorithm} presents the details of UnbiasedGBM. From the algorithm, we can see that $\mathrm{score}_2$ determines the feature to split, and $\mathrm{score}_3$ is the unbiased gain that determines whether to perform leaf-wise early stopping. 

\begin{figure}[ht]
\centering
    \subfigure[10 medium-scale datasets with only numerical features.]{
        \begin{minipage}[t]{0.95\linewidth}
        \centering
        \includegraphics[width=\textwidth]{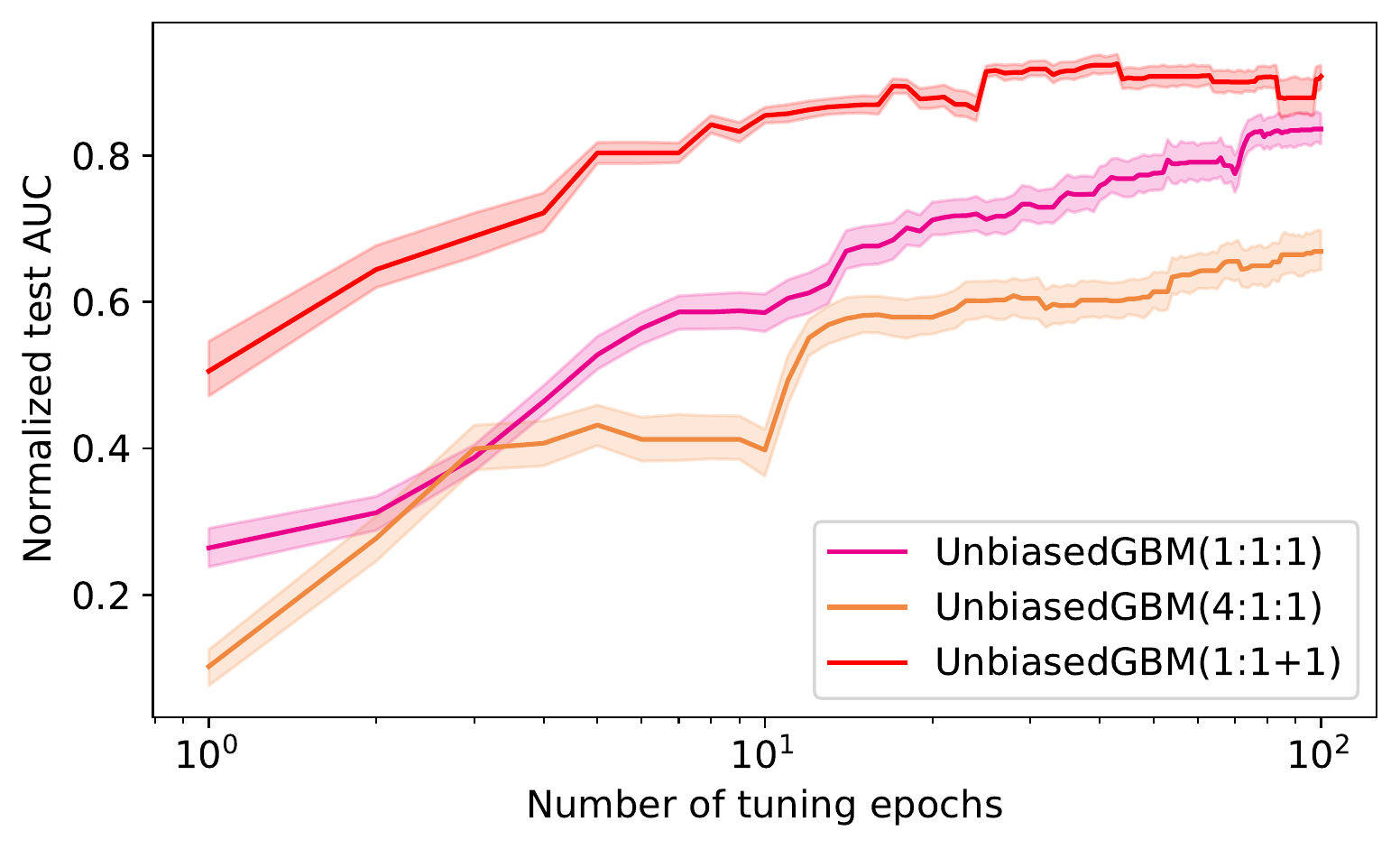}
        \end{minipage}%
    }
    
    \subfigure[10 medium-scale datasets with num. and cat. features.]{
        \begin{minipage}[t]{0.95\linewidth}
        \centering
        \includegraphics[width=\textwidth]{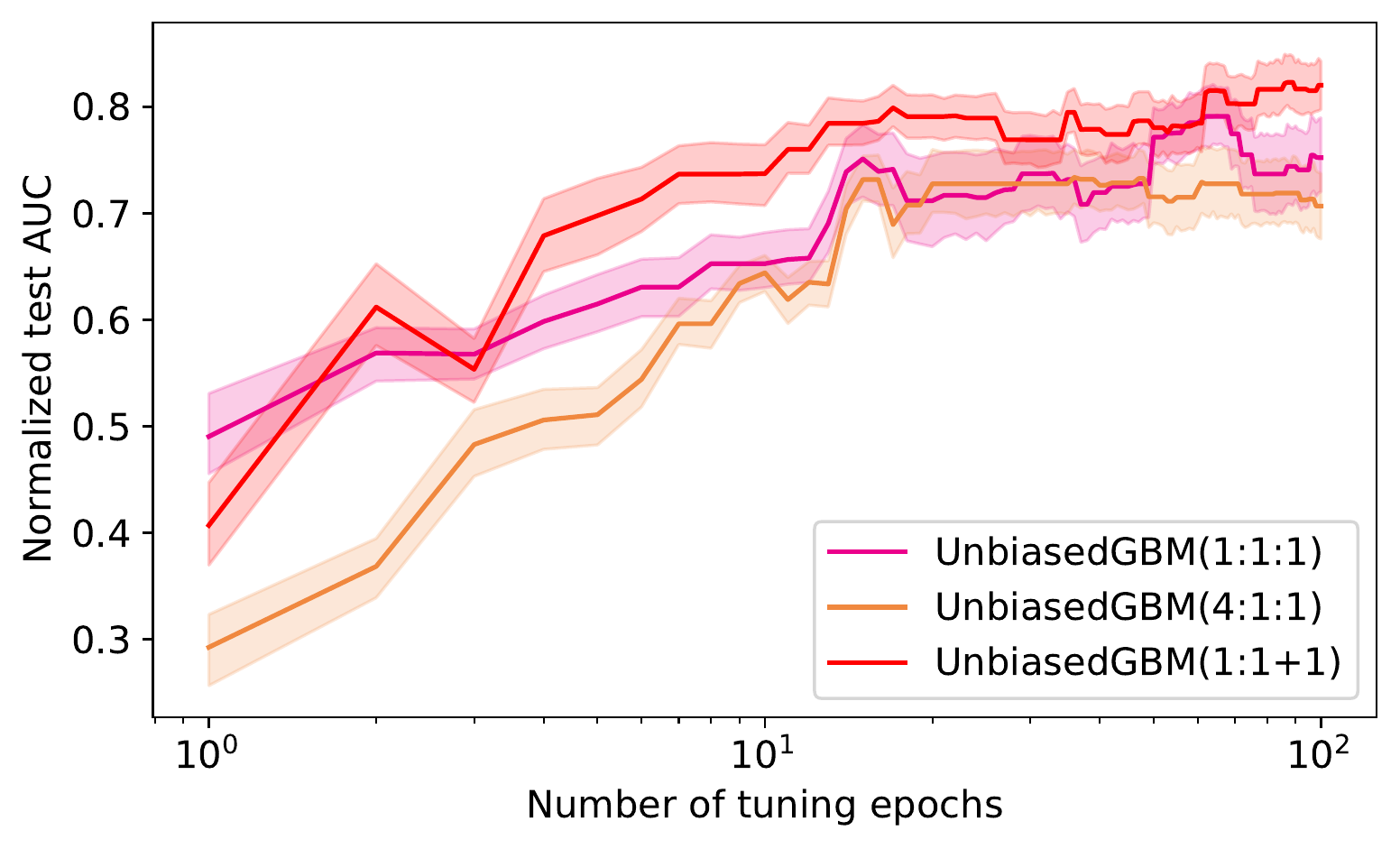}
        \end{minipage}
    }%

\caption{We investigate the influence of the proportion of splitting data. We find that splitting the dataset evenly ($|\mathcal D|:|\mathcal D_1'|:|\mathcal D_2'|=1:1:1$) is more reasonable than unevenly ($4:1:1$). We can further improve the performance by setting $\mathcal D_1' = \mathcal D_2'$ (1:1+1). Each value corresponds the normalized test AUC of the best model (on the validation set) after a specific number of tuning epochs, averaged on all the datasets. The shaded area presents the variance of the scores.}
\label{fig: appen}
\end{figure}

In fact, $\mathrm{score}_2$ is nearly unbiased when the number of features is small. As a result, for the sake of sample efficiency, we can set $\mathcal D_1' = \mathcal D_2'$ in the applications, and thus $\mathrm{score}_2 = \mathrm{score}_3$. We find that such a design is beneficial to the performance in our main experiments.
We present the results in Figure~\ref{fig: appen}.


\begin{algorithm}[t]
    \caption{Split Finding}
    \label{alg:algorithm}
    \textbf{Input}: $S_I$, instance set of current node\\
    \textbf{Output}: The best split with its gain
    
    \begin{algorithmic}[1] 
\STATE Randomly seperate $S_I$ into $S^{(1)}$, $S^{(2)}$, $S^{(3)}$.
\STATE Let $b_{i}$ indicate instance $i$ in $S^{(b_i)}$
\STATE $G^{(k)} \leftarrow \sum_{i\in S_k} g_k$, $H^{(k)} \leftarrow \sum_{i\in S_k} g_k$
\STATE $\mathrm{score}^{(2)}_{1,2,3}, \theta^{(2)} \leftarrow (-\inf,-\inf, -\inf), \mathrm{None}$
\FOR{$f$ in feature space}
    \STATE $G^L_{1,2,3} \leftarrow 0,0,0$, $H^L_{1,2,3} \leftarrow 0,0,0$
    \STATE $\mathrm{score}^{(1)}_{1,2,3}, \theta^{(1)} \leftarrow (-\inf,-\inf, -\inf), \mathrm{None}$
    \FOR{$i$ in sorted($S$, ascent order by $\mathbf x_{if}$)}
        \STATE $G^L_{b_i} \leftarrow G^L_{b_i}+g_i$, $H^L_{b_i} \leftarrow H^L_{b_i}+h_i$
        \STATE $G^R_{b_i} \leftarrow G_{b_i}-G^L_{b_i}$, $H^R_{b_i} \leftarrow H_{b_i}-H^L_{b_i}$
        \STATE $\mathrm{score}_1 \leftarrow \frac{G^L_1G^L_1}{H^L_1} + \frac{G^R_1G^R_1}{H^R_1} -\frac{G_1G_1}{H_1}$
        \STATE $\mathrm{score}_2 \leftarrow \frac{G^L_1G^L_2}{H^L_2} + \frac{G^R_1G^R_2}{H^R_2} -\frac{G_1G_2}{H_2}$
        \STATE $\mathrm{score}_3 \leftarrow \frac{(G^L_{1}+G^L_{2})G^L_3}{H^L_3} + \frac{(G^R_{1}+G^R_{2})G^R_3}{H^R_3} -\frac{(G_{1}+G_{2})G_3}{H_3}$
        \IF{$\mathrm{score}_1 > \mathrm{score}_1^{(1)}$} 
            \STATE $\mathrm{score}^{(1)}, \theta^{(1)} \leftarrow \mathrm{score}, (f,\mathbf x_{if})$
        \ENDIF
    \ENDFOR
    \IF{$\mathrm{score}_2^{(1)} > \mathrm{score}_2^{(2)}$} 
        \STATE $\mathrm{score}^{(2)}, \theta^{(2)} \leftarrow \mathrm{score}^{(1)}, \theta^{(1)}$
    \ENDIF
\ENDFOR
\RETURN $\mathrm{score}^{(2)}_3, \theta^{(2)}$
    \end{algorithmic}
\end{algorithm}

\subsection{Tree Construction}

When constructing a decision tree, we repeatedly split the leaf with maximal score. Algorithm~\ref{alg:algorithm2} shows the details.

\subsection{Time Complexity}

Let $n$ be the number of samples, $m$ be the number of base features of dataset. Each sample appears exactly once of each depth, so with maximum depth $d$, our implementation runs in 
$$
O\left(Tdnm\log n\right)
$$ 
where $T$ is the number of trees. This complexity is exactly the same as XGBoost and similarly cost $O(Tdnm + nm\log n)$ on the block structure. In fact, applying our method to existing GBDT implementations preserves their time complexity, because it is never worse than calculating on $2$ more separated dataset $D_1'$ and $D_2'$.

\begin{algorithm}[!t]
    \caption{Tree Construction}
    \label{alg:algorithm2}
    \textbf{Output}: Decision tree with feature importance gain
    
    \begin{algorithmic}[1]
        \STATE $T \leftarrow$ a root only
        \STATE $\mathrm{Imp} \leftarrow [0,0,0,...]$
        \WHILE {$|T| < \mathrm{num\_leaf}$}
            \STATE Pick the leaf $I$ with maximal $\mathrm{Split}_{\mathrm{score}}$
            \STATE $\mathrm{Imp}[\mathrm{Split}_{\mathrm{feat}}]\ += \mathrm{Split}_{\mathrm{score}}$
            \IF{$\mathrm{Split}_{\mathrm{score}} < \mathrm{min\_split\_gain}$}
                \STATE \textbf{break}
            \ENDIF
            \STATE $T \leftarrow T \cup \{I \mapsto I_L, I_R\}$
        \ENDWHILE
        \RETURN $T$, $\mathrm{Imp}$
    \end{algorithmic}
\end{algorithm}

\paragraph{Discussion: Complexity when applied on XGBoost.}

XGBoost sorts all instances for each feature when determining the best split on a node. The bottleneck is to visit the sorted array of instance once and calculate its split gain. In this case, using our method incurs no additional costs because the total number of instances of $\mathcal D$, $\mathcal D_1'$, and $\mathcal D_2'$ equals to the original.

\paragraph{Discussion: Complexity when applied on LightGBM.} 

LightGBM divides instances into bins. When the number of bins is not small, the bottleneck is to visit each sorted bins and calculate its split gain. If we separate $\mathcal D$, $\mathcal D_1'$ and $\mathcal D_2'$ over the bins, the total number of bins of the three dataset is the same as the original.
Hence no additional costs again.

\begin{table}[]
    \centering
    \resizebox{\linewidth}{!}{
    \begin{tabular}{ccccc} \toprule
          & SHAP & Permutation & Gain Importance & Unbiased Gain \\ \midrule
         Average Rank & 3.14 & 2.71 & 2.71 & 1.43 \\ 
         p-value & $0.003$ & $0.042$ & $0.042$ & - \\ \bottomrule
    \end{tabular}
    }
    \caption{The average rank over 14 datasets and the p-value of Nemenyi test between unbiased gain and the baseline methods.}
    \label{tab: average rank feat select}
    
\end{table}

\section{Additional Results}
\subsection{Statistical Analyses}
We leverage the Nemenyi test~\cite{DBLP:journals/jmlr/Demsar06} to compare the unbiased gain and baseline methods in feature selection. For each dataset, we average the AUC on the test set when selecting top $10\%$, $20\%$, and $30\%$ features. We present the rank of each method and the Nemenyi test p-values in Table \ref{tab: average rank}.

 \begin{figure}[ht]
\centering
    \subfigure[14 high-dimensional datasets]{
        \begin{minipage}[t]{0.48\linewidth}
        \centering
        \includegraphics[width=\textwidth]{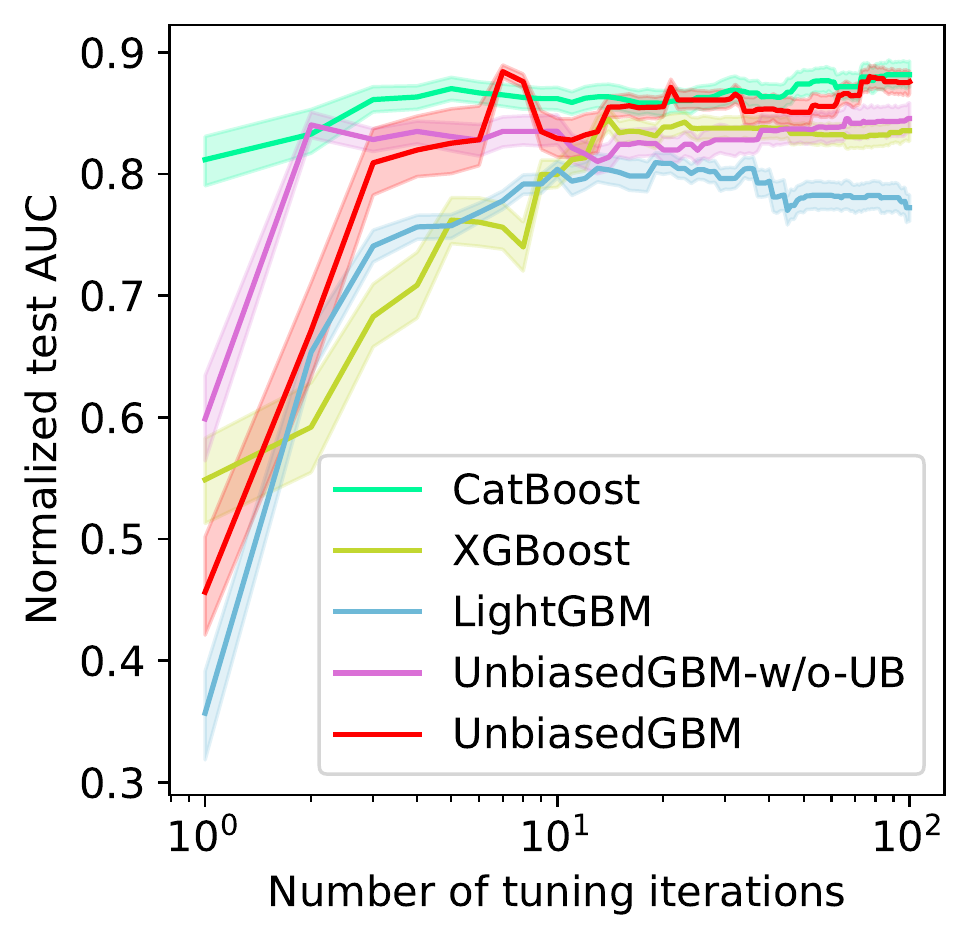}
        \end{minipage}%
    }%
    \subfigure[20 easy datasets]{
        \begin{minipage}[t]{0.48\linewidth}
        \centering
        \includegraphics[width=\textwidth]{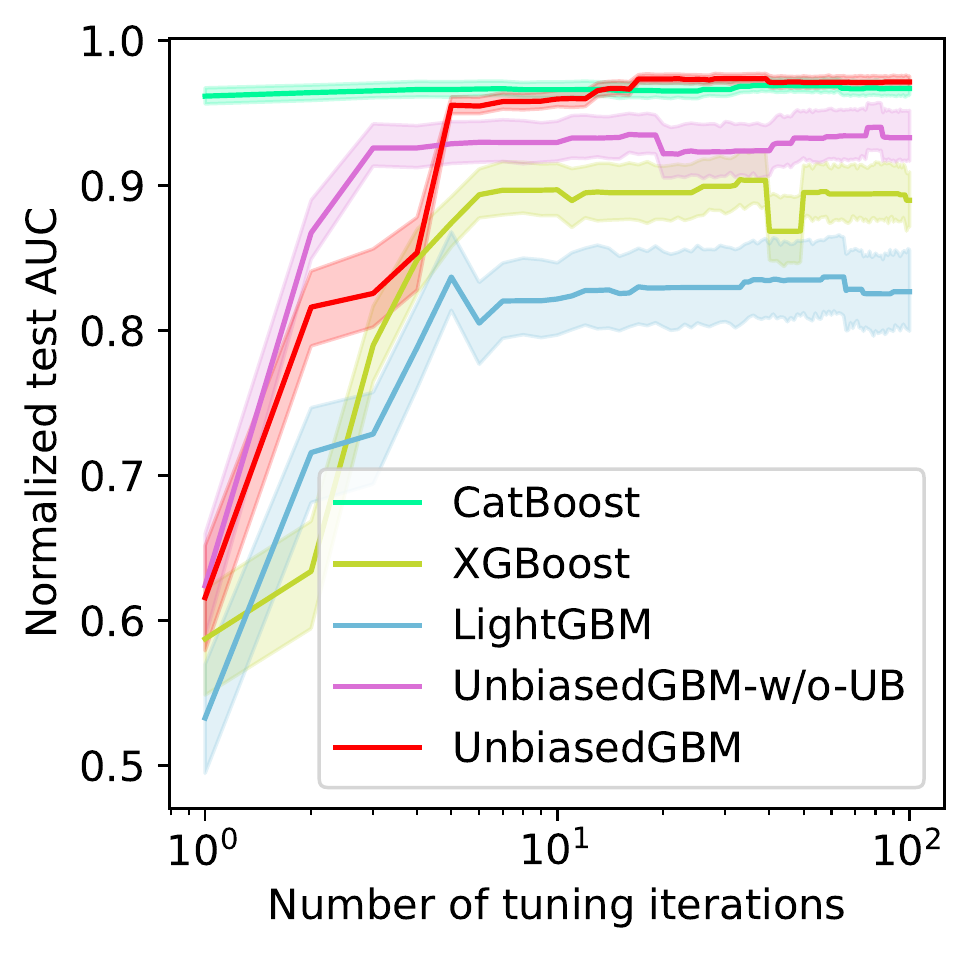}
        \end{minipage}%
    }%
    
\caption{Additional experiments of high dimensional and easy datasets.}
\label{app: fig: additional_experiments}
\end{figure}

\subsection{Additional Experiments}
We present additional experiments on high dimensional and easy datasets in Figure \ref{app: fig: additional_experiments}.

\end{document}